\documentclass[10pt,authoryear]{elsarticle}

\usepackage[utf8]{inputenc}
\usepackage[authoryear]{natbib}
\usepackage{graphicx}
\usepackage[margin=0.8in]{geometry}
\usepackage{hhline} 
\usepackage{array} 
\usepackage[table]{xcolor} 
\usepackage{amsmath}
\usepackage{amssymb}
\usepackage{color}
\usepackage{siunitx}
\usepackage{pdflscape}

\usepackage{setspace}

\journal{Remote Sensing of Environment}

\newcommand{\beginsupplement}{
	\setcounter{table}{0}
	\renewcommand{\thetable}{S\arabic{table}}
	\setcounter{figure}{0}
	\renewcommand{\thefigure}{S\arabic{figure}}
}

\begin{document}
	
	\begin{frontmatter}
		
		\title{SLIC-UAV: A Method for monitoring recovery in tropical restoration projects through identification of signature species using UAVs}
		
		\author[a,b]{Jonathan Williams\corref{*}}
		\ead{jonvw28@gmail.com}
		\author[b]{Carola-Bibiane Sch\"onlieb}
		\author[a,c]{Tom Swinfield}
		\author[d]{Bambang Irawan}
		\author[d]{Eva Achmad}
		\author[d]{Muhammad Zudhi}
		\author[e]{Habibi}
		\author[e]{Elva Gemita}
		\author[a]{David A. Coomes\corref{*}}
		\ead{dac18@cam.ac.uk}
		
		\address[a]{Department of Plant Sciences, University of Cambridge Conservation Research Institute, David Attenborough Building, Cambridge, CB2 3QY, UK}
		\address[b]{Image Analysis Group, Department of Applied Mathematics and Theoretical Physics, University of Cambridge, CB3 0WA, UK}
		\address[c]{Centre for Conservation Science, Royal Society for Protection of Birds, David Attenborough Building, Cambridge, CB2 3QY, UK}
		\address[d]{Faculty of Agriculture, Jambi University, Jambi, Indonesia}
		\address[e]{PT Restorasi Ekosistem Indonesia, Jl. Dadali No. 32, Bogor 16161, Indonesia}
		\cortext[*]{Corresponding authors}

		\date{June 2020}
		
		\begin{abstract}
			\label{sec:abs}
			Logged forests cover four million square kilometres of the tropics, capturing carbon more rapidly than temperate forests and harbouring rich biodiversity. Restoring these forests is essential if we are to avoid the worst impacts of climate change. Yet monitoring tropical forest recovery is challenging. Tracking the abundance of visually identifiable, early-successional species enables successional status and thereby restoration progress to be evaluated. Here we present a new pipeline, SLIC-UAV, for processing Unmanned Aerial Vehicle (UAV) imagery to map early-successional species in tropical forests. The pipeline is novel because it comprises: (a) a time-efficient approach for labelling crowns from UAV imagery for training and validating models; (b) machine learning of species based on spectral and textural features within individual tree crowns, and (c) automatic segmentation of orthomosaiced UAV imagery into 'superpixels', typically smaller than tree crowns, using Simple Linear Iterative Clustering (SLIC) to delineate regions. Creating these superpixels massively reduces the dataset's dimensionality and focuses prediction onto clusters of pixels of the same species, greatly improving accuracy. To demonstrate SLIC-UAV, support vector machines and random forests were used to predict the species of hand-labelled crowns in a restoration concession in Indonesia. The random forest approach proved most accurate at discriminating species from the background, with accuracies ranging from 79.3\% when mapping five common species, to 90.5\% when mapping the three most visually-distinctive species. In contrast, support vector machines provided the better modelling approach for labelling automatically segmented superpixels, with accuracies ranging from 74.3\% to 91.7\% for the same species labels. Predictive models were then extended to map species across 100 hectares of forest. The study demonstrates the utility of SLIC-UAV for mapping characteristic early-successional tree species as an indicator of successional stage within tropical forest restoration areas. We show how species densities can be used to produce dominance maps within management units, enabling forest condition to be easily discerned across restoration projects. Continued effort is needed to develop easy-to-implement and low-cost technology to improve the affordability of project management.
		\end{abstract}
		
		
	\end{frontmatter}
	
	\section{Introduction}
	\label{sec:intro}
	
	Tropical forest restoration is central to ameliorating the worst impacts of global climate breakdown while simultaneously protecting vast swathes of terrestrial biodiversity \citep{Palmer2008,Duffy2009,Thompson2009,Isbell2011,Myers2000,Joppa2011,Bastin2019}. The IPCC have called for``unprecedented changes in all aspects of society", including reversing the forecast loss of 2.5 million $\textrm{km}^2$ of forest to a 9.5 million $\textrm{km}^2$ increase in forest cover by 2050 \citep{IPCC2018}. Logged-over tropical forests are particularly important carbon sinks because they are widespread, covering 4 million $\textrm{km}^2$ \citep{Cerullo2019}, and capture carbon rapidly as they recover lost biomass \citep{Edwards2014}. Natural tropical forests, such as these, are more likely to be successfully restored and persist \citep{Crouzeilles2017}, and host vastly more biodiversity value than actively managed forests \citep{Edwards2014}. But natural tropical forests continue to be threatened by agricultural expansion and the intention of many countries to use fast-growing plantations to meet international restoration commitments is a serious concern \citep{Lewis2019}. It is therefore of critical importance to develop remote sensing methods capable of assessing restoration performance in terms of biodiversity recovery to complement the already advanced techniques for measuring above-ground biomass \citep{Asner2010,Aerts2011,Melo2013,Chave2014,Zahawi2015,Iglhaut2019}.
	
	Biodiversity recovery may correlate poorly with above-ground biomass in regenerating tropical forests, and measuring species richness is often intractable due to the thousands of species involved, so developing reliable indicators of biodiversity is necessary \citep{Martin2015,Sullivan2017}. Although biodiversity typically increases as forests accumulate above-ground biomass, the relationship is complicated by disturbance history, fragmentation and active management, so that forests of equivalent biomass harbour very different levels of biodiversity \citep{Slik2002,Slik2008,Sullivan2017}. For example, a single round of logging removing 100$\textrm{m}^3$ of wood per hectare may result in a 40\% reduction in biomass but only a 10\% reduction in biodiversity \citep{Martin2015}, whereas a fast-growing plantation can rapidly accumulate biomass without a corresponding increase in biodiversity \citep{Bernal2018}. To properly account the benefits of forest restoration it is therefore important to assess biodiversity, but tropical forests may host more than 1,000 species per hectare \citep{Myers2000,Joppa2011}. This makes direct measurements of species richness in restoration projects prohibitively costly using field measurements potentially hampering direct estimation by remote sensing \citep{Turner2003,Sullivan2017}. Instead it may be possible to assess biodiversity by assessing the abundance and composition of early-successional species: following disturbance, early-successional species including grasses, shrubs, lianas, and fast-growing trees become abundant, often representing more than 30\% of the canopy \citep{Slik2002,Slik2003b,Slik2008}. These species have adaptations that make them competitive in high light environments, including large, thin leaves (e.g. low leaf mass per area), long petioles, open canopies and high foliar nutrient concentrations, which also make them visually distinct and easy to identify \citep{Slik2009}. If disturbance ceases, early-successional species gradually become less frequent, through their mortality and failure to recruit in the shaded-understory, making them valuable indicators of both historic disturbance and subsequent recovery \citep{Slik2003a,Slik2008}.
	
	Quantifying recovery of secondary tropical forest in terms of indicative early-successional species still requires methods which scale to enable cost-effective application across management units and remote sensing approaches are able to offer this \citep{Petrou2015,Fassnacht2016,Almeida2019}. Traditional approaches to biodiversity or species occurrence monitoring rely on field observations that sample only only a tiny fraction of the landscape \citep{Turner2003}, which is also true for newer approaches, including environmental DNA and functional trait measurements, each allowing diversity to be viewed from a different lens \citep{Asner2009,Zhang2016,Bush2017,Colkesen2018}. Remotely-sensed satellite imagery can be used to interpolate data from field plots based upon variability of spectral signatures, estimating variation and approximating species composition across landscapes \citep{Adelabu2013}, but the spatial and temporal resolution of most satellite imagery remains a constraint \citep{Carleer2004}, and higher resolution data, such as those collected from aircraft, are needed to monitor individual trees \citep{Bergseng2015}. By combining of aerial laser scans with hyperspectral or multispectral imagery species can be mapped \citep{Zhang2012,Alonzo2013,Dalponte2014,Maschler2018,Marconi2019}, with crown-level precision if the resolution of the sensors is sufficient \citep{Ballanti2016,Fassnacht2016}. However, these sensors are often custom-designed or prohibitively expensive where commercially available, which limits the accessibility of these surveys \citep{Surovy2019}. Finding a balance between feasibility, cost and utility is key to seeing methods adopted and approaches must adapt to emerging technologies \citep{Turner2003,Toth2016,Kitzes2019}.
	
	Unmanned Aerial Vehicles (UAVs) offer a cheap remote sensing methodology which increases the temporal and spatial resolution of imagery available and are increasingly adopted in forest research \citep{Saari2011, Anderson2013, Bergseng2015,Rokhmana2015, Surovy2019}. UAVs are being deployed to map insect damage \citep{Nasi2015}, post-logging stumps \citep{Samiappan2017}, flowering events \citep{Lopez-Granados2019}, leaf phenology \citep{Park2019}, and forest biomass \citep{Dandois2013,Zahawi2015} but analytical methods to evaluate forest recovery in terms of species composition or biodiversity with UAVs are lacking \citep{Messinger2016,Goodbody2018a}. Even approaches to detect tree species from UAV imagery remain scarce and are often limited to predicting species for manually delineated crowns \citep{Lisein2015,Tuominen2018} or else work in other ecological contexts such as high latitude \citep{Puliti2017,Alonzo2018,Franklin2018}, riparian strips \citep{Michez2016} or managed nurseries \citep{Gini2018}. All of these methods require manual field data collection, either in the form of complete plot inventory \citep{Puliti2017} or crown delineation with GPS\citep{Alonzo2018}, taking time and access to trees, which is tricky in the tropics and approaches to collect reference data should take advantage of new technologies to improve efficiency. Further, detailed mapping of species across management units from UAV imagery requires methods that can extend knowledge of species for a sample of crowns to a whole region. Object-based image analysis on UAV imagery offers promise for mapping tree species in this way, allowing use of textural information computed over regions of adjacent pixels, rather than simply evaluating the pixel values individually \citep{,Giannetti2018,Gini2018,Lu2019,Puliti2019}. Typically, regions of interest are manually-defined such as pre-defined management units, inventory plots or tree crowns, for which statistics are generated \citep{Lisein2015,Michez2016,Alonzo2018,Franklin2018,Tuominen2018}, meaning models can only apply to other similarly created objects. Extending these approaches to all imagery across a site requires automated partitioning of imagery into groups of neighbouring pixels (superpixels) which act as the objects \citep{Ren2003}. Superpixels labelled with species identities are used to build and validate models that can then be applied to all superpixels, covering the whole landscape \citep{Feduck2018,Wu2019}. This approach has yielded promising results in limited settings, such as conifer seedling mapping in logged 50 $\textrm{m}^2$ plots and mapping a single invasive species across an island in japan \citep{Feduck2018}, but the approach has not been applied to detect early-successional species in recovering tropical forests. Developing and applying UAV technologies to tropical forest restoration settings to map key species indicative of disturbance and recovery trajectory will help improve efficiency of management of forest restoration: knowing where interventions are most likely to work or most needed can reduce labour costs \citep{Rose2015}.
	
	\paragraph{Our contribution:} This study presents, UAV-SLIC, a novel and complete workflow for mapping early-successional species in degraded tropical forests. We developed this end-to-end pipeline combining UAV data collection with an object-based approach to learn species from the textural and spectral properties of superpixel clusters, enabling extension of data from a sample of crowns to map indicative species occurrence across 130 ha of forest. In contrast to existing methods, UAV-SLIC enables wall-to-wall mapping of multiple early-successional species, so that forest recovery and successional status can be evaluated. The UAVs we used are rapidly deployed, commercially available, and can map approximately 100 ha per day, enabling small sites to be mapped in their entirety, large sites to be sampled, and repeat surveys to track recovery through time. We evaluate the performance of conventional red-green-blue (RGB) and multispectral (RGB+NIR; using a \$3,000 camera) imagery, comparing the accuracy of both types of data. We develop an integrated UAV-based approach to collect species identity data, greatly reducing time and effort in the field, and use oil palm to show that additional categories can be added through desk-based mapping. We reveal that object-based image recognition from low-cost UAV imagery is highly accurate for detecting early-successional species against a background of typical forest species, which has the potential to transform forest restoration evaluations. Finally, we produced heat-maps across 100 ha of forest to reveal the spatial signature of disturbance, created by logging, which can be used as a baseline for tracking recovery and directing restoration management.
	
	\paragraph{Outline of the paper:} We introduce the methodology underlying our SLIC-UAV approach, including generating superpixels with SLIC and extracting spectral and textural features for each superpixel as well as how we used UAVs to efficiently collect reference data. We then compare models for labelling early-successional tree crown imagery from Hutan Harapan forest in Indonesia based on lasso regression, support vector machines (SVM) and random forests (RF). We validate the feature extraction and labelling by first using manually delineated crowns, before exploring the fully automated SLIC-UAV pipeline by mapping crown labels onto superpixels, finding the SVM model to be best. We then use this modelling approach to extend our species labelling to a region of 100 ha of forest at Hutan Harapan. We find the method has particular strength with identifying early-successional species particularly noted for being indicative of recency and severity of disturbance, showing the power of this UAV-based mapping approach for aiding forest restoration management.
	
	\section{Materials and Methods}
	\label{sec:meth}
	
	In this section we introduce the SLIC-UAV method, explaining the steps of superpixel extraction, feature generation and subsequent predictive modelling comparing three options: Lasso Regression, Support Vector Machines and Random Forests. We also introduce our study data used to illustrate the use of SLIC-UAV, including the development of a data collection pipeline using a UAV in place of traditional field survey to reduce the time needed to curate a reference set of crowns used for predictive modelling of key species of interest, focusing on early-successional species indicative of disturbance and typical long-lived early-successional species.
	
	\subsection{Data Collection}
	\label{ssec:meth_data}
	
	\subsubsection{Study Site}
	\label{sssec:meth_data_hh}
	
\begin{figure}[ht]
		\begin{center}
			\includegraphics[width=\textwidth]{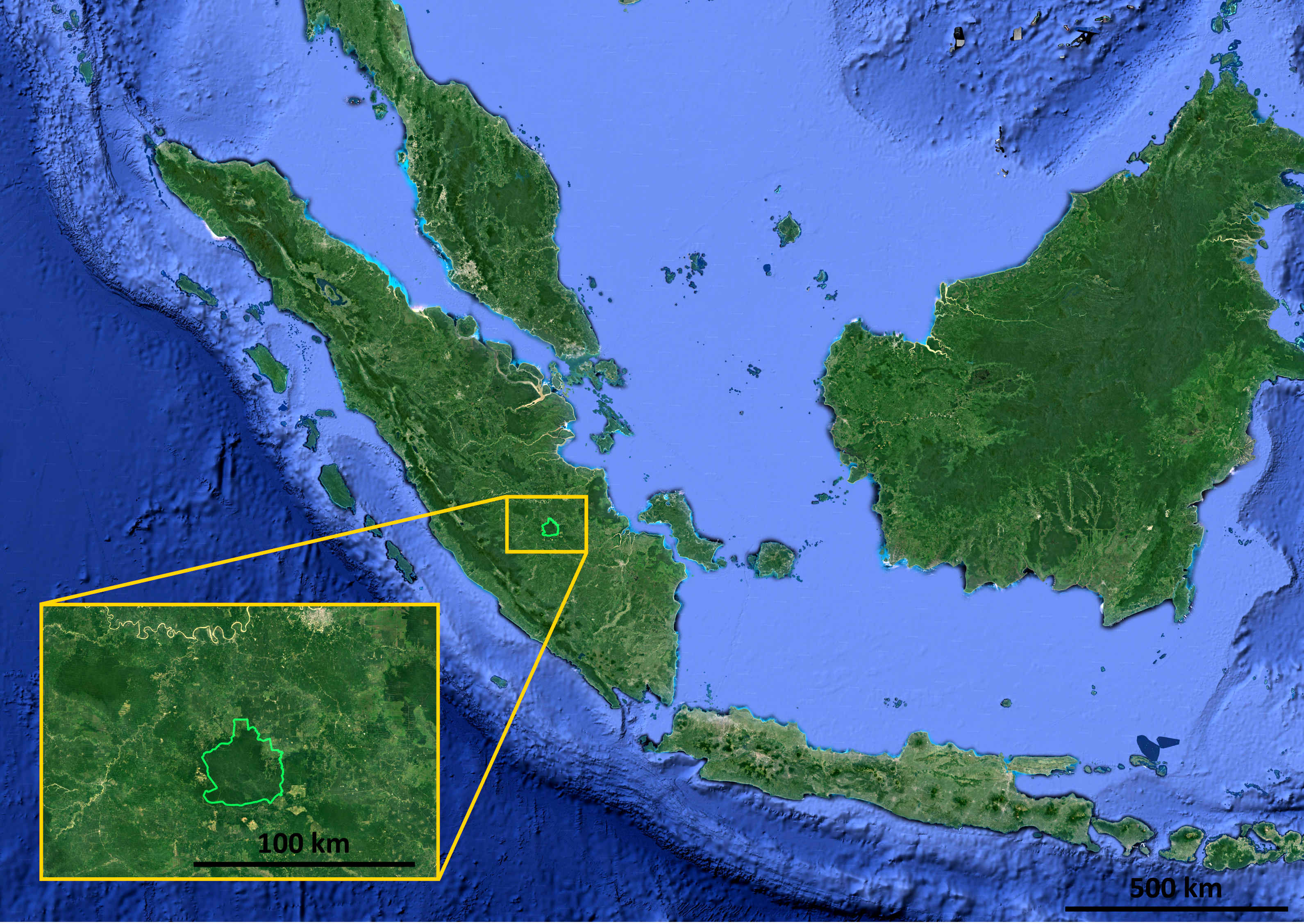}
			\caption{Location of Hutan Harapan within Indonesia (green polygon). Imagery data courtesy of Google}
			\label{fig:harapan}
		\end{center}
	\end{figure}

	Data for this study were collected at Hutan Harapan (``forest of hope'') on the island of Sumatra, Indonesia (Figure \ref{fig:harapan}). Hutan Harapan is an Ecosystem Restoration Concession where 98,455ha of ex-logging concessions are now leased for restoration \citep{Harrison2015}. Heavy logging occurred since the 1970s, resulting in a heterogeneous secondary lowland dipterocarp forest in various stages of recovery. Harapan has a weakly seasonal climate with monthly mean rainfall varying from 79 mm to 285 mm, with a dry season with less than 100 mm of rain for three consecutive months between June and August. The terrain at Harapan in undulating, however elevation remain low in the range 30-120 m above sea level. Despite heavy logging since the 1970s, Harapan supports a large amount of biodiversity, with 302 bird species and over 600 tree species from 107 plant families recorded \citep{Harrison2015}.
	
	Our study site comprised a 130 ha area close at the boundary of Hutan Harapan West of the main camp. The site was characterised by a closed-canopy forest with pre-disturbance remnant trees emerging from dense regrowth of early-successional trees including \textit{Macaranga spp.} (Euphorbiaceae) and the invasive pioneer \textit{Bellucia pentamera} (Melastomataceae) from South America \citep{deKok2015}. These species are common in across the entire landscape and within disturbed forest more generally in Southeast Asia \citep{Slik2003a,Slik2008,Dillis2018}. The study area also included oil palm within the adjacent concession. Data were collected in two survey periods in 2017 and 2018 as we now outline.

	\subsubsection{UAV Imagery}
	\label{sssec:meth_data_imagery}
	
	Multispectral (MS) data were collected in April 2017. For this a 3DR Solo UAV (3DR, Berkeley, USA) was equipped with a Parrot Sequoia (Parrot, Paris, France) camera held in a fixed mount angled close to nadir when flying at mission speed. The camera records standard red-green-blue (RGB) imagery at 16 megapixel resolution as well as four bands of multispectral imagery. For this study use only the multispectral data. These fall into four bands with centres and approximate response widths (both in nm): Red (550, 40), Green (660, 40), Red Edge (735, 10) and Near Infrared (790, 40). Images in these bands are recorded at 1.2 megapixel resolution, giving a ground sampling distance of 14.8 cm per pixel at an altitude of 120 m. Additionally, a sensor atop the UAV records illumination in each band at the time of exposure allowing radiometric correction of illumination to reflectance values, reducing the effect of varying solar illumination. UAV flights were flown by autopilot. Each flight covered a 10.75 ha footprint in a grid designed in QGIS \citep{QGIS}. Mission Planner \citep{MissionPlanner} was used to design the flight path, in a snaking pattern with 80\% in-line and 70\% between-line overlap between images, also referred to as front-lap and side-lap. 
	
	Standard digital RGB data for this study were collected in November 2018. For this a DJI Phantom 4 UAV was used with its stock camera (DJI, Hong Kong, China). This camera records standard RGB imagery at 12.4 megapixel resolution, which at a height of 100 m gives a pixel size of 4.35 cm. UAV flights were flown by autopilot. Each flight covered a 10.75 ha footprint in a grid designed in QGIS. The DJI GS Pro app was used to plan each flight, in a snaking pattern with 80\% in-line and 70\% between-line overlap, all flown at an altitude of 100 m. Data were processed as detailed below. We collected these additional data as the RGB sensor on the Parrot Sequoia uses a rolling shutter, leading to blurring of images making them insufficient quality for further processing. The multispectral imagery did not include the blue band and we also wished to compare the contribution of raw RGB imagery, as typical from a cheaper consumer camera, to that of the corrected finely tuned multispectral data for species discrimination.
	
\begin{figure}[ht]
		\begin{center}
			\includegraphics[width=\textwidth]{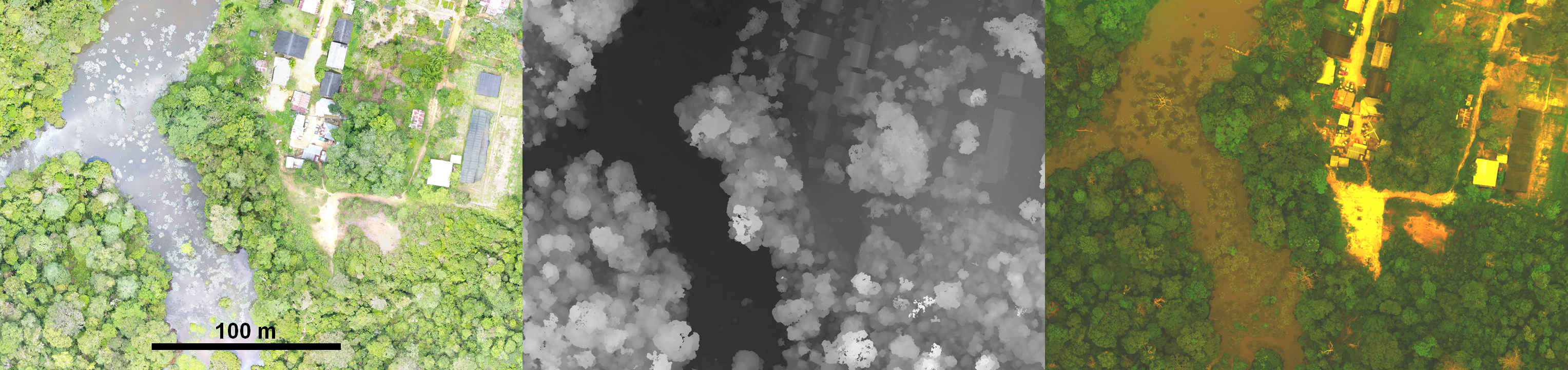}
			\caption{Examples of the three imagery types used in this work, standard Red-Green-Blue, Digital Surface Model and Multispectral (left to right). The multispectral imagery is a false-colour image with red, green and blue representing red, green and red edge channels.}
			\label{fig:imagery}
		\end{center}
	\end{figure}

	Agisoft Photoscan Professional (Agisoft, St. Petersburg, Russia) was used to process both datasets. For the multispectral data the steps were to align photos, calibrate reflectance (using data from the sunshine sensor to correct for illumination based upon the data from the onboard sensor which is known to improve reliability of classification methods \citep{Tuominen2018}), build a dense cloud, build a DEM (digital elevation model) and build an orthomosaic. This allowed the algorithm to refine estimates of camera location from initial GPS stamps, correct for illumination, build a photogrammetric model for the area and extract outputs of a rasterised DEM and orthomosaic (OM) of the image mosaics after correcting for the surface geometry. Parameters for each step are listed in Table \ref{table:param}. For the MS imagery this process took approximately 34 hours using a workstation running Windows 7 equipped with an Intel Xeon E3-1240 V2 CPU, comprising 8 cores running at 3.4 GHz with 16 GB RAM, though this included producing a dense point cloud. Had we worked with only the sparse option, this process would take 14 h 35 m. RGB imagery was treated in the same way, without reflectance calibration. The higher resolution of the RGB data required the process be separated into chunks to enable loading into the 16 GB of RAM on the desktop used, taking a total of approximately 11 days to run, though sticking to the sparse point reduced this to 17 h 59 m. We chose to use the dense cloud for higher structural detail, but all steps could be completed with only the sparse cloud. Splitting the data used overlapping chunks which were then aligned and merged using the built-in marker-based alignment. Markers were manually set for all flights with overlap between the chunks, with a minimum of 4 clearly visible fixed locations used for each pair of flights with overlap, taking roughly one hour of human input. The RGB and MS data co-aligned using the Georeferencer tool in QGIS (version 3.4.5). Tie points were generated as a random set of 100 points across the region. We ensured good correspondence using clearly visible features present in both datasets close to each random point as the final tie point. A further 51 points were then plotted by starting from the centroid of the largest remaining voronoi polygons across the network of tie points. Drawing tie points took roughly four hours of human input. These points were then used to transform the MS data to align with the RGB data using a polynomial transformation (polynomial 3 in QGIS) with nearest neighbour resampling. The final outputs used in this study were a multispectral orthomosaic (MS), RGB-derived surface elevation model (DSM) and RGB orthomosaic (RGB) with pixel resolutions of 11.3 cm, 8.01 cm and 4.01 cm respectively (examples shown in Figure \ref{fig:imagery}).
	
	\begin{table}[h]
		\caption{Values of Parameters used in Agisoft Photoscan}
		\begin{center}
			\begin{tabular}{|>{\centering\arraybackslash}m{11em}|c|>{\centering\arraybackslash}m{8em}|c|>{\centering\arraybackslash}p{7em}|>{\centering\arraybackslash}p{6em}|}
				\hline
				\multicolumn{2}{|c|}{Align Photos} & \multicolumn{2}{|c|}{Build Dense Cloud} & \multicolumn{2}{|c|}{Build DEM} \\
				\hline
				Parameter & Value & Parameter & Value & Parameter & Value \\
				\hhline{|-|-|-|-|-|-|}
				Accuracy & High & Quality & High & Source Data &  Dense Cloud\\
				Adaptive Camera Model Fitting & Off & Calculate Point Colours & On & Interpolation & Enabled \\
				\hhline{|~|~|~|~|-|-|}
				Generic Pre-selection & On &  Depth Filtering & Aggressive & \multicolumn{2}{|c|}{Build Orthomosaic}\\
				\hhline{|~|~|~|~|-|-|}
				Reference Pre-selection & On & && Parameter & Value  \\
				\hhline{|~|~|~|~|-|-|}
				Key Point Limit & 40,000 & & & Surface & DEM \\
				Tie Point Limit & 8,000 & & & Blending Mode & Mosaic \\
				\hline
			\end{tabular}
		\end{center}
		\label{table:param}
	\end{table}
	
	\subsubsection{Generating labelled tree crowns for training}
	\label{sssec:meth_data_tree_loc}
	
\begin{figure}[h]
		\begin{center}
			\includegraphics[width=\textwidth]{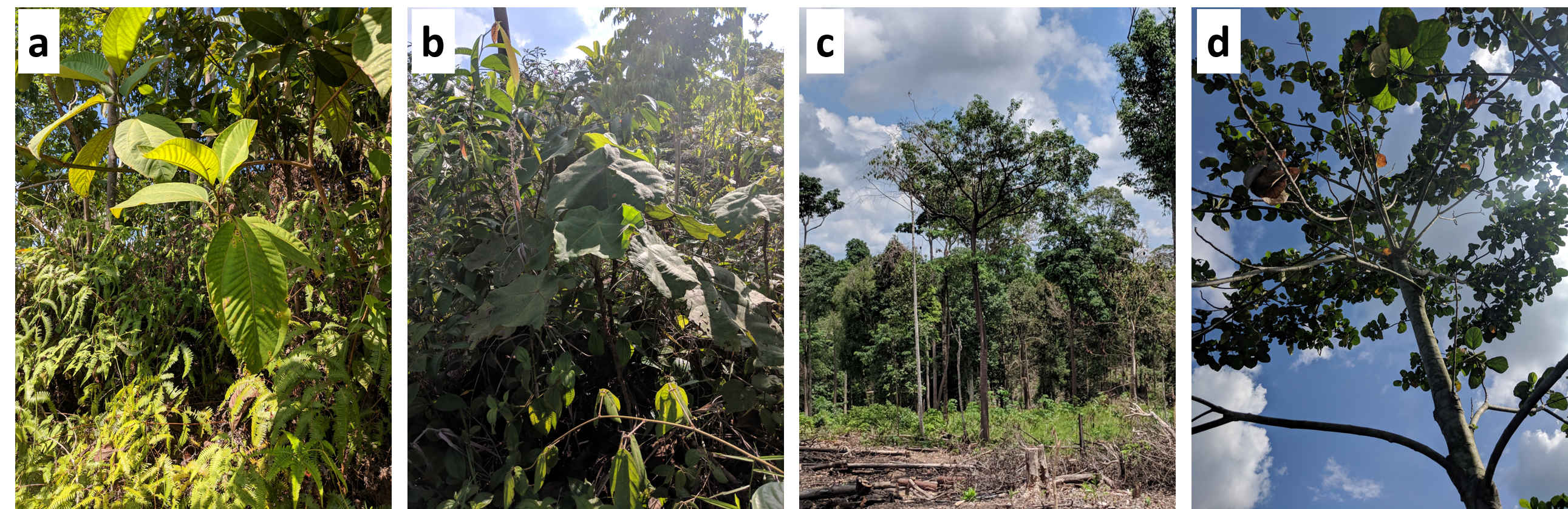}
			\caption{Individuals of the four main species as photographed in Harapan: a) \textit{Bellucia pentamera}, b) \textit{Macaranga gigantea}, c) \textit{Alstonia scholaris} and d) \textit{Endospermum malaccense}}
			\label{fig:tree_ex}
		\end{center}
	\end{figure}

	A set of georeferrenced hand-drawn polygons were produced for the four species of interest in November 2018. We focused on two early-successional species and two species indicative of more established secondary forest. The early-successional species were \textit{Macaranga gigantea} (Euphorbiaceae) and the non-native invasive \textit{Bellucia pentamera} (Melastomataceae) which are both prevalent across Harapan, especially in more degraded areas. For we also chose two long-lived early-successional species, with less visually distinct crown and leaf traits: \textit{Alstonia scholaris} (Apocynaceae) is a long-lived tree that can reach to 60 m in height and produces commercially valuable timber; \textit{Endospermum malaccense} (Euphorbiaceae), locally known as Sendok-sendok, is a mid-canopy tree, reaching 34 m in height typical of secondary regrowth \citep{Slik2009}. Examples of these species are shown in Figure \ref{fig:tree_ex}. Visually the `long-lived early-successional' species appear more similar, both from the ground and from above, making identification of these from each other, and other upper canopy trees, difficult. The earlier successional species are easy to spot from the ground owing to their low height and distinctive leaves. Their leaf arrangements also lead to striking patterns and textures when visualised from above, making these easier to identify by eye from UAV imagery.

	\begin{figure}[ht]
		\begin{center}
			\includegraphics[width=0.8\textwidth]{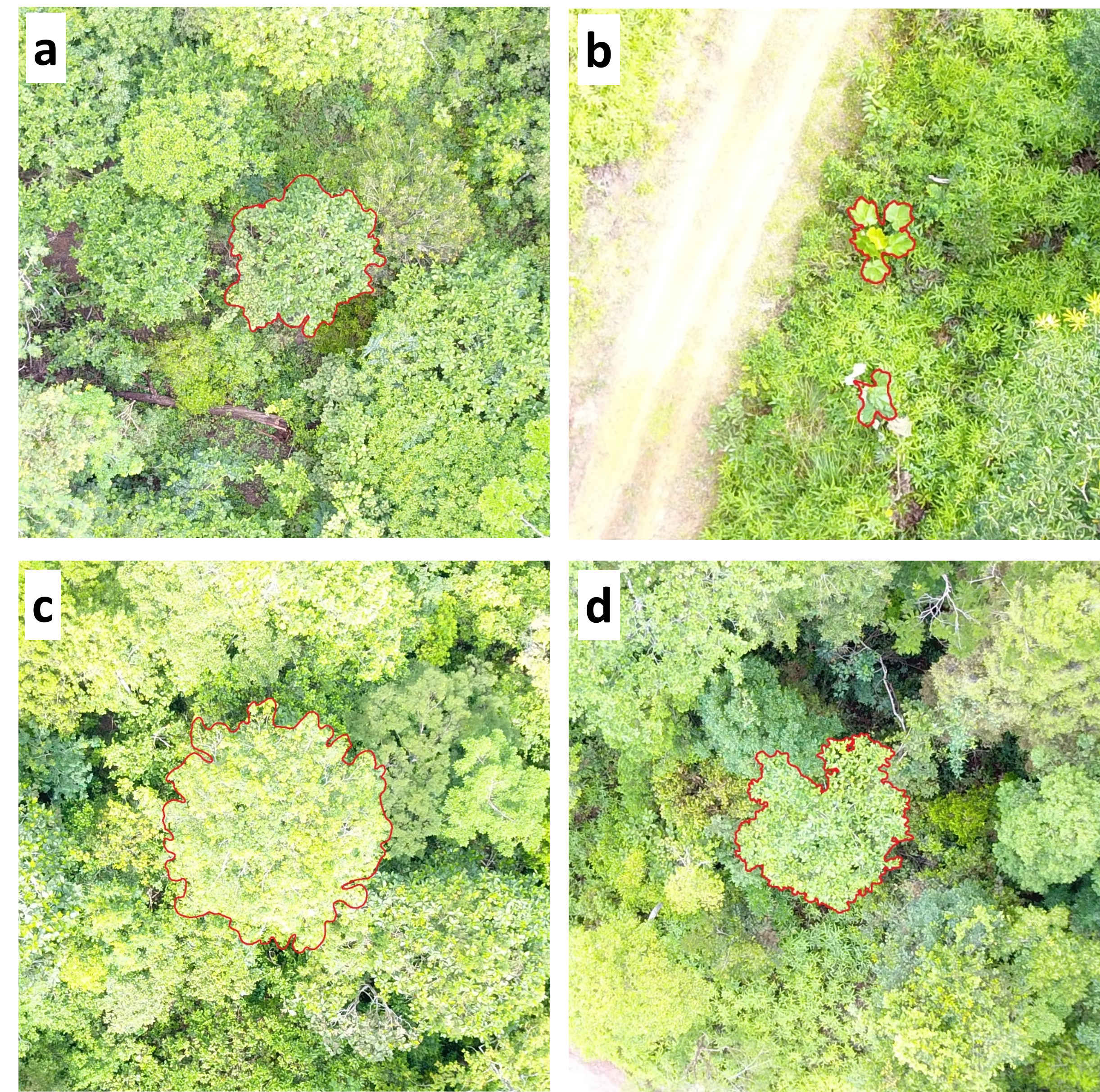}
			\caption{Example manual digitisation of the four species mapped in the field: a) \textit{Bellucia pentamera}, b) \textit{Macaranga gigantea}, c) \textit{Alstonia scholaris} and d) \textit{Endospermum malaccense}}
			\label{fig:drawn_crown}
		\end{center}
	\end{figure}
	
	We used the Phantom 4 UAV to record the location of trees, which has two significant advantages over the traditional approach of mapping trees from the forest floor: (1) above-canopy UAV-based GPS measurements are high precision (\textless$\pm$3 m) when compared with sub-canopy hand-held GPS measurements ($\pm$15 m); and (2) high quality images of tree crowns are collected that facilitate the production of hand-drawn tree crown polygons. Our application is focused on mapping just four key species across whole management units and so we chose this alternative to traditional exhaustive surveying of small plots to obtain data for building our model. We worked along roads and at other locations from which we could launch the UAV and then scanned the canopy at low altitude (typically 20-30 m above the canopy) using the UAV flown in manual mode to identify crowns and species from the live high resolution imagery; images were reviewed by local experts to confirm identifications. Once we were happy we had identified a crown of interest, two team members would position themselves at right angles to each other relative to the crown centre. The UAV operator would then move the UAV horizontally until both team members agreed it was above the crown. Here multiple images were captured at various heights. Images from about 30 crowns were collected in this way in a morning, and manually digitised in the afternoon (while the trees were fresh in memories). The crown boundaries were marked on the highest resolution image with reference to images at multiple heights, and any crowns with unclear boundaries were re-confirmed in the next batch of flights. Once RGB OM and DSM rasters were produced from mapping surveys, the crowns were converted to geospatial polygons using the initial annotated images and GPS tags. The raw imagery was used as the primary reference, whilst using contrast in the RGB OM and boundaries in height in the DSM to refine any boundaries. In total, data were collected for 328 crowns: \textit{B. pentamera} (n = 120), \textit{M. gigantea} (n = 65), \textit{A. scholaris} (n = 93) and \textit{E. malaccense} (n = 50); example digitisations shown in Figure \ref{fig:drawn_crown}.
	
\begin{figure}[ht]
		\begin{center}
			\includegraphics[width=0.8\textwidth]{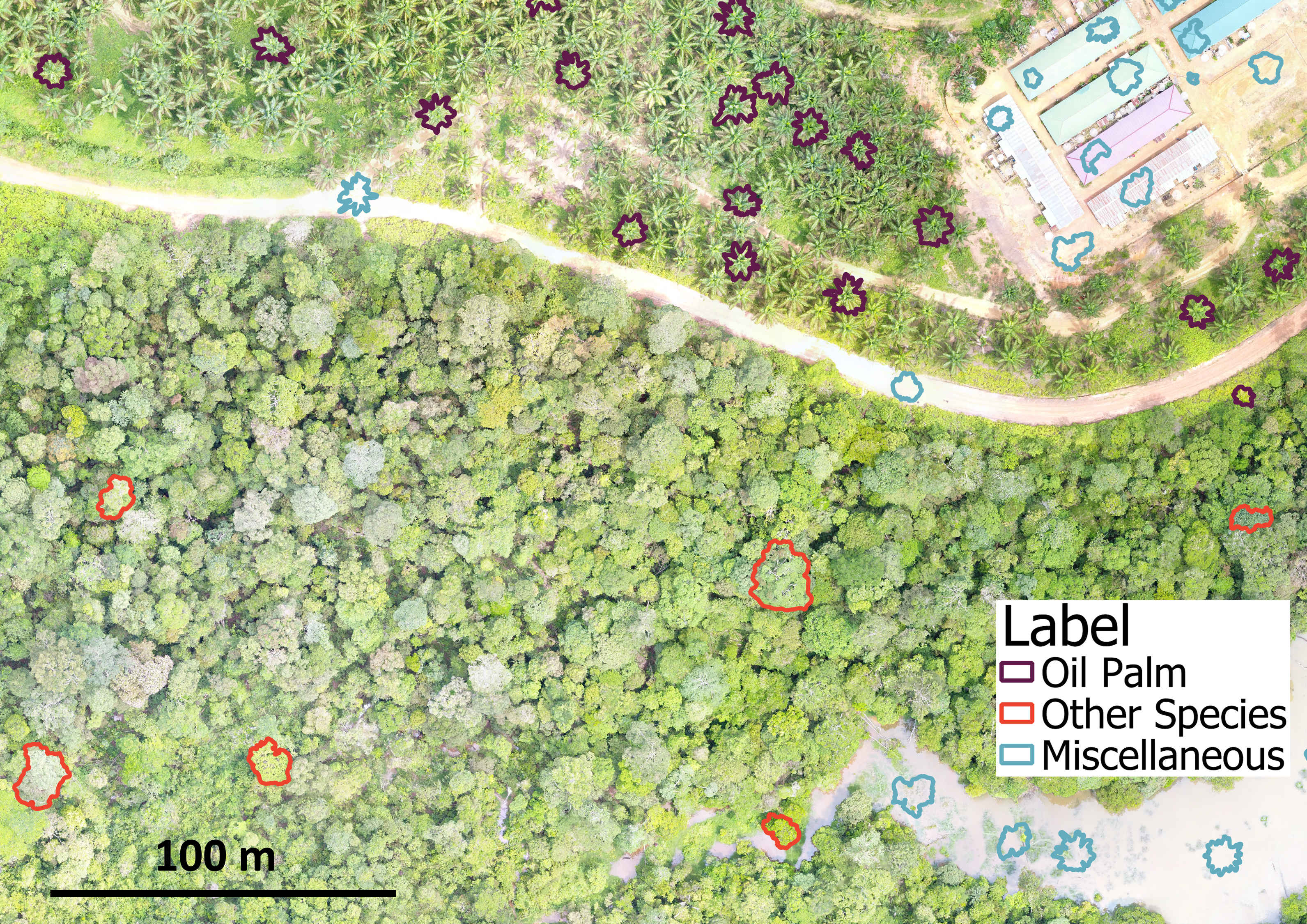}
			\caption{Example manual digitisation of oil palm, other species and miscellaneous regions completed with UAV imagery}
			\label{fig:extra_crown}
		\end{center}
	\end{figure}

	In addition to the crowns mapped in the field, we also digitally delineated three other classes to include in our model, using all three forms of imagery overlaid to confidently extract crown extent. We digitised 105 oil palm crowns, drawn at random locations across the plantations or (occasionally) within the recovering forest. We also digitised 100 `other' crowns by drawing points at random across our study site. We considered each of these carefully, checking the original (higher resolution) imagery to ensure each crown was not in fact one of the four target species. Finally, in order to map the complete area, we took 100 existing crown outlines of varying species, size and structure and placed these over non-vegetated regions (including water bodies, roads, bare ground and buildings) to allow our models to distinguish the miscellaneous non-vegetation regions present in the data. We used existing outlines from the other labels to ensure similar sizes for each polygon, as well as an even balance of area for the classes. Examples of the new labels are shown in Figure \ref{fig:extra_crown}. Overall this gave us 633 labelled regions across seven categories: five focal species, one other tree class and a miscellaneous class.
	
	\subsection{Identifying Species with UAV-SLIC}
	\label{ssec:meth_algo}
	
	SLIC-UAV centres on an object-based image analysis (OBIA) workflow \footnote{available at https://github.com/jonvw28/SLICUAV}. This enables the context and patterns of imagery to be used in modelling, in contrast to a pixel-based approach which focuses only on the local spectral data for each pixel \citep{Yu2006,Colkesen2018}. First regions are defined either manually (hand-drawn crown polygons) or automatically (using image segmentation as detailed in Section \ref{sssec:meth_algo_segs}). Next we extract features for the imagery data in each region that are used as input to predictive species labelling models; we trialled lasso regression, support vector machines (SVM) and random forests (RF). We first test these approaches using manually delineated crowns before we extend this approach to automated region delineation, creating superpixels (clusters of neighbouring pixels) through simple linear iterative clustering (SLIC) using the RGB imagery data from 2018. We apply the same pipeline as for manually delineated crowns to build predictive models for these. We assess this approach on manually-delineated crowns and on the subset of automatically generated image segments that intersected the manually-delineated crowns, yielding two validation approaches. Finally, after validation, the trained models were used to predict species labels for automatically-created segments across the 100 ha study site, eventually building heatmaps of canopy dominance (proportion of area predicted as each species) to indicate forest condition.
	
	\subsubsection{Superpixel Segmentation with SLIC}
	\label{sssec:meth_algo_segs}
	
	Within our approach, we learn and predict species for superpixel objects, combining many individual pixels to compare then in their local context. Superpixel segmentation separates imagery into disjoint groups of connected pixels, based on similarity \citep{Ren2003}. This can be partial or full segmentation. Partial segmentation extracts an incomplete set of pixels from the image. In our case this applies when we manually delineate the crowns we mapped from the imagery and only these extracted regions are used in models. Complete segmentation instead assigns every pixel to a group, called a superpixel, which can be thought of as a partition of the imagery into segments. We use this approach in our automated landscape mapping, enabling extension of the pipeline to complete coverage of any region where imagery exists.
	
\begin{figure}[ht]
		\begin{center}
			\includegraphics[width=0.8\textwidth]{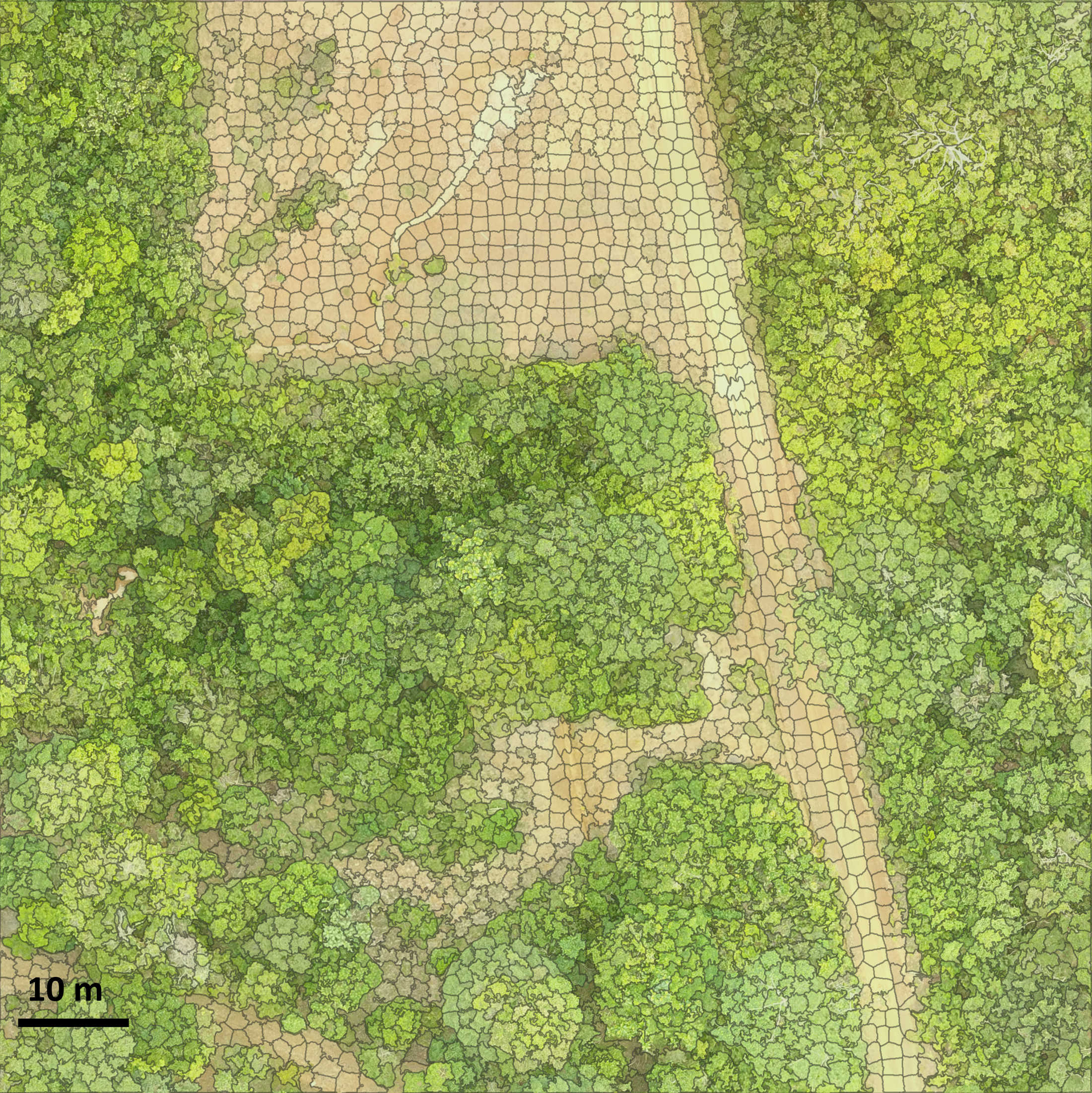}
			\caption{Example region segmentation into superpixels using SLIC for 1 ha of imagery at Harapan}
			\label{fig:SLIC}
		\end{center}
	\end{figure}

	Automated region segmentation was completed within SLIC-UAV, using the RGB imagery from 2018, by Simple Linear Iterative Clustering (SLIC) \citep{Achanta2010,Achanta2012}, see Figure \ref{fig:SLIC} for an example. This is similar to k-means clustering, but is designed to produce regions of roughly similar area in a regular spacing by starting with a regular grid of squares. These are then iterated using k-means clustering in a local neighbourhood, four times the average superpixel size, using a weighted sum of euclidean distance between pixel locations and distance in colour space as the distance metric. Once superpixel centres become sufficiently stable (based on sequential changes) connectivity is enforced, ensuring all pixels in a given superpixel are locally connected. The algorithm adapts to contours of the image, like k-means, but the regularity constraints ensure superpixels have a similar size. The size is then mostly controlled by the number of initial superpixels. For our work we used the implementation of SLIC in the scikit-image Python library, using Python 3.7 \citep{walt2014,Python}. We used the default compactness of 10 and sigma of 1, and initialised superpixels to have an average area of 0.5 $\mathrm{m}^2$ to ensure these were smaller than all but the smallest crowns.
	
	\subsubsection{Feature Extraction}
	\label{sssec:meth_algo_feat}
	
	\begin{figure}[h]
		\begin{center}
			\includegraphics[width=\textwidth]{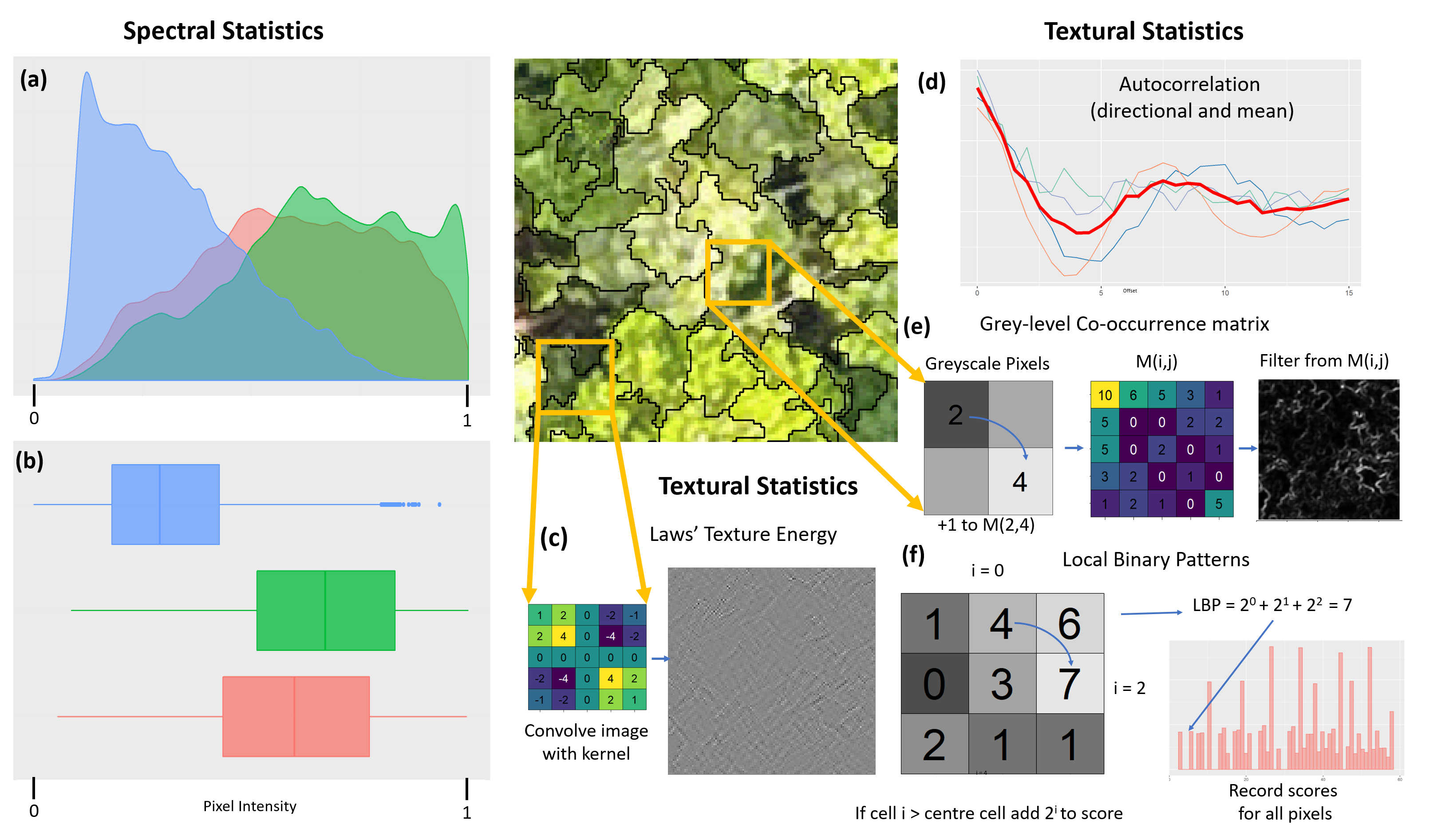}
			\caption{Illustration of the key features used in this work. On the left are example visualisations of distribution for parametric (a) and non-parametric statistics (b) for RGB pixels in the central image. (c) is an example of a filter used for Laws\textsc{\char13} textural features, with the E5E5 kernel for detecting edges. (d) shows autocorrelation scores in four directions spaced by 45\si{\degree} with average score overlaid. (e) is an example calculation of a grey-level co-occurrence matrix and the filter given by the dissimilarity measure in a local 7 $\times$ 7 neighbourhood, (f) is an example of computing a Local Binary Pattern, with a histogram of occurrences of each motif score across the inset imagery. }
			\label{fig:meth_feats_examp}
		\end{center}
	\end{figure}
	
	Imagery for each region (crown or superpixel) was used to generate a set of summary features to use for species prediction. We generated features for each of the three imagery types, treating the DSM as a greyscale raster with floating point values, which were all scaled and centred to mean zero and variance one based on training data. We computed features in two broad classes: spectral and textural. Spectral features are based on summary statistics of the individual pixel values in the imagery, as is commonly used in UAV mapping approaches \citep{Ota2015,Kachamba2016}. In contrast, textural features were computed by treating the superpixel as an image, computing statistics based on repeating patterns and frequencies of pattern motifs in the arrangement of pixels \citep{Franklin2018}. Example visualisations of these concepts are shown in Figure \ref{fig:meth_feats_examp}. Vegetation indices were computed as stated in Table \ref{table:spec_ind} based on the bands of the orthomosaics and treated as extra spectral bands for the spectral analysis such as in \citet{FuentesP2018} and \citet{Goodbody2018b}. Similarly, we converted the RGB imagery into HSV space treating hue, saturation and value as additional spectral bands \citep{Smith1978}. In an effort to focus on only illuminated portions of each superpixel, we also filtered the top 50\% brightest pixels, defined by lightness in CIELAB colour space \citep{CIELAB1976}, and computed the RGB spectral features for just these pixels. We computed the same statistics on the RGB bands, the MS bands, the DSM float imagery, the RGB and MS indices, the HSV channels and the bands of the brightness filtered RGB imagery. Details of all the spectral statistics computed for each the data types are listed in Table \ref{table:spec_feat}.
	
	Textural features were produced from four approaches: the greylevel co-occurrence matrix (GLCM) \citep{Haralick1973}, local binary patterns (LBP) \citep{He1990,Ojala1996}, Laws\textsc{\char13} features \citep{Laws1980} and spatial autocorrelation. RGB data were transformed into a greyscale image for all textural features and each of the four multispectral bands were treated as a greyscale image and had textural features computed independently, whilst DSM imagery was treated in different ways depending on the texture method. GLCM statistics summarise patterns or frequently co-occurring local pairs of pixel values, with both the mean and range of scores when considering all directions with the distance of offset reported. For the DSM data we converted the float values to 32 integer values, defined as a linear spacing (and rounding) from 1 for heights below the 5th percentile within that region to 32 for heights above the 95th percentile. GLCM statistics were computed for offsets of 1, 2 and 3 pixels. LBPs quantifies the frequency of patterns of relative pixel values for a neighbourhood of a given radius. We used a rotationally invariant form of LBPs counting all motifs equivalent up to a rotation as one single pattern. The DSM was treated the same way as for GLCM, and again this was applied at radii of 1, 2 and 3 pixels. Laws\textsc{\char13} features compute convolutions of the imagery with particular kernels constructed as a cross product of vectors designed to identify spots, waves, lines, ripples and intensity. Imagery is first modified by subtracting the mean value in a 15x15 window for each pixel. Then each 5x5 kernel is convolved with the resulting image and we report the mean and standard deviation for resulting pixel values, using the float version of DSM imagery in this case. Spatial autocorrelation scores the correlation of the image with itself, and we recorded the mean correlation across all directions for each offset, along with the range across all directions. We computed this for 1, 2 and 3 pixel offsets, again using float data when looking at the DSM. Details of all textural features are listed in Table \ref{table:text_feat}.
	
	\subsubsection{Predicting Species}
	\label{sssec:meth_algo_crown}
	
	We considered three main approaches to predicting the species within the SLIC-UAV pipeline for each region based upon the features we extracted. Our first approach was to use lasso regression which gives a sparse model that can readily be interpreted \citep{Tibshirani1996}. Our second approach was to apply a support vector machine (SVM) to the features, being a less restrictive but harder to interpret model \citep{Cortes1995}. Finally both approaches were compared to random forests modelling \citep{Ho1995,Ho1998}. For all approaches we assessed the classifiers with 10-fold cross-validation. Ten models were fitted on 90\% of the crowns, with the remaining 10\% used for validation. Folds were split in a random stratified way, to balance each species label equally across all folds, with the test sets forming a complete covering of all crowns where each crown was used to build nine of the models and test the tenth, independently built model. This split was the same for all models. For automated region mapping, we added labels to all regions with 50\% or greater of their area within a labelled crown, leading to multiple labelled superpixels for most crowns. In total this produced 11,996 manually labelled superpixels. The splitting of the training and test superpixels in each fold of cross-validation was also based on the original crown split, keeping all superpixels for a given crown in the same set to avoid inflation of accuracy from training on superpixels within crowns which are included in the test set.
	
	Lasso regression (LR) is an extension to least-squares regression which regularises the coefficients of the resulting model. This both reduces the likelihood that the model is heavily reliant on any one feature, but more critically, restricts the number of predictors included in the model. We fitted models using the \texttt{glmnet} package in R \citep{R}. Here we used a multinomial logistic regression to give relative confidence scores for each class label, with the highest species being the final prediction. Data examples were weighted inversely proportionally to the number of examples with that species label to account for mismatching number. We constructed our models to ensure that where a feature was included for one class, it was included for all classes, and restricted our models to be the best-fitting model (based on overall accuracy on the training set) which had at most 25 features included in the model.
	
	SVM modelling was completed using the \texttt{e1071} package in R. Here the model was fitted to the training data using the default radial basis function kernel with parameters tuned by the inbuilt method and class weights set to balance the contribution of each class, as for lasso regression. The model was allowed to use all variables in contrast to the restriction applied in lasso regression. Similarly, the random forest models were built on the training data, using the R package \texttt{randomForest}, again applying weights to correct for varying class size. Here 500 trees were used, with the default tree structure used (sampling $\textrm{sqrt}(f)$ variables at each node, where $f$ features are supplied to the model). All approaches used here have a built in within-sample validation for model parameter selection.
	
	We explored the contribution of different data sources and feature types for the reliability of our models within SLIC-UAV. We considered the three data sources (RGB, MS and DSM) and two features classes (spectral and textural) separately. We explored the effect of using all valid combinations of data sources combined with using either or both classes of features. In total this gave us 21 possible model input options in addition to the model using all variables. Given the number of replicates, we chose to use a single training and test data split, using 75\% of crowns to train each model and 25\% to evaluate, keeping the split the same for all combinations.
	
	As noted previously, visually the difference between the long-lived early-successional species, \textit{Endospermum malaccense} and \textit{Alstonia scholaris}, and other canopy species was subtle. This was in contrast to \textit{Bellucia pentamera} and \textit{Macaranga gigantea}, with distinctive structure and leaf texture, whose occurrence is known to be closely related to disturbance history. We also noted that we had a small sample of \textit{E. malaccense} crowns and that these were often more difficult to confidently identify in the field. We therefore considered modelling where these trees were included with the other tree species category, keeping only \textit{Alstonia scholaris} as an indicative long-lived early-successional species example. Finally, we also considered models where this was included with other crowns to form a `lower management concern' class, actively seeking only the invasive \textit{Bellucia pentamera}, early-successional \textit{Macaranga gigantea} and potentially encroaching palm oil tree classes compared to species more indicative of progress beyond initial stages of succession. We believe this final model, whilst simpler, is potentially of more interest to forest managers.
	
	\subsection{Landscape Mapping}
	\label{ssec:meth_land}

	\begin{figure}[ht]
		\begin{center}
			\includegraphics[width=0.9\textwidth]{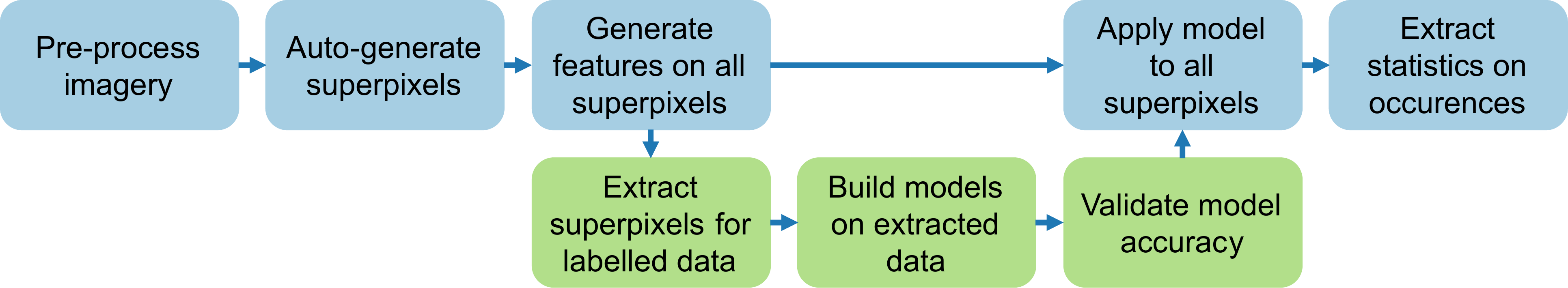}
			\caption{Schematic of the full SLIC-UAV pipeline. Blue steps are applied to the full imagery, whereas green steps only apply to superpixels for which manual class labels are transferred from the manually annotated dataset}
			\label{fig:meth_pipeline}
		\end{center}
	\end{figure}
	
	We applied our SLIC-UAV superpixel approach to produce labels across the whole study site for which we had imagery. For this we used selected the best performing model, and then retrained a classifier given all training data. This produced maps of species occurrence, which were used to compute the area and percentage of cover of each species. We also used these to produce density maps by computing the percentage prevalence of each species in a grid, where each cell was 0.25 ha in size. As there was a region of clearance to establish agroforestry between the UAV surveys, we masked regions where this had occurred by the 2018 survey when computing the landscape models of species occurrence, since the multispectral data would include trees no longer present. This doesn\textsc{\char13}t affect the model building phases: of the labelled crowns only six `other' species trees were in this region, all of which were in small fragments of canopy left after clearing and were verified to be of very similar extent in both years of imagery, both in the processed imagery and original source images. We expect the outputs from models across the study site to be of direct value for measuring successional status and directing active restoration towards areas highlighted as being early-successional or dominated by low biodiversity value species. The full pipeline for this approach is summarised in Figure \ref{fig:meth_pipeline}.


	\section{Results}
	\label{sec:res}
	
	We validated our SLIC-UAV approach, first by computing features and predictive models using the manually digitised crown. We then assessed the performance of our approach for the full SLIC-UAV pipeline, automatically generating superpixels and building models on these. We then proceeded with our best model to extend our labels for the mapped crowns to the whole of the 100 ha landscape we worked with at Hutan Harapan. Finally we investigated the relative effects of the imagery and features we used on mapping performance.
	
	\subsection{Crown-level Species Models}
	\label{ssec:res_crown}
	
	\begin{figure}[ht]
		\begin{center}
			\includegraphics[width=\textwidth]{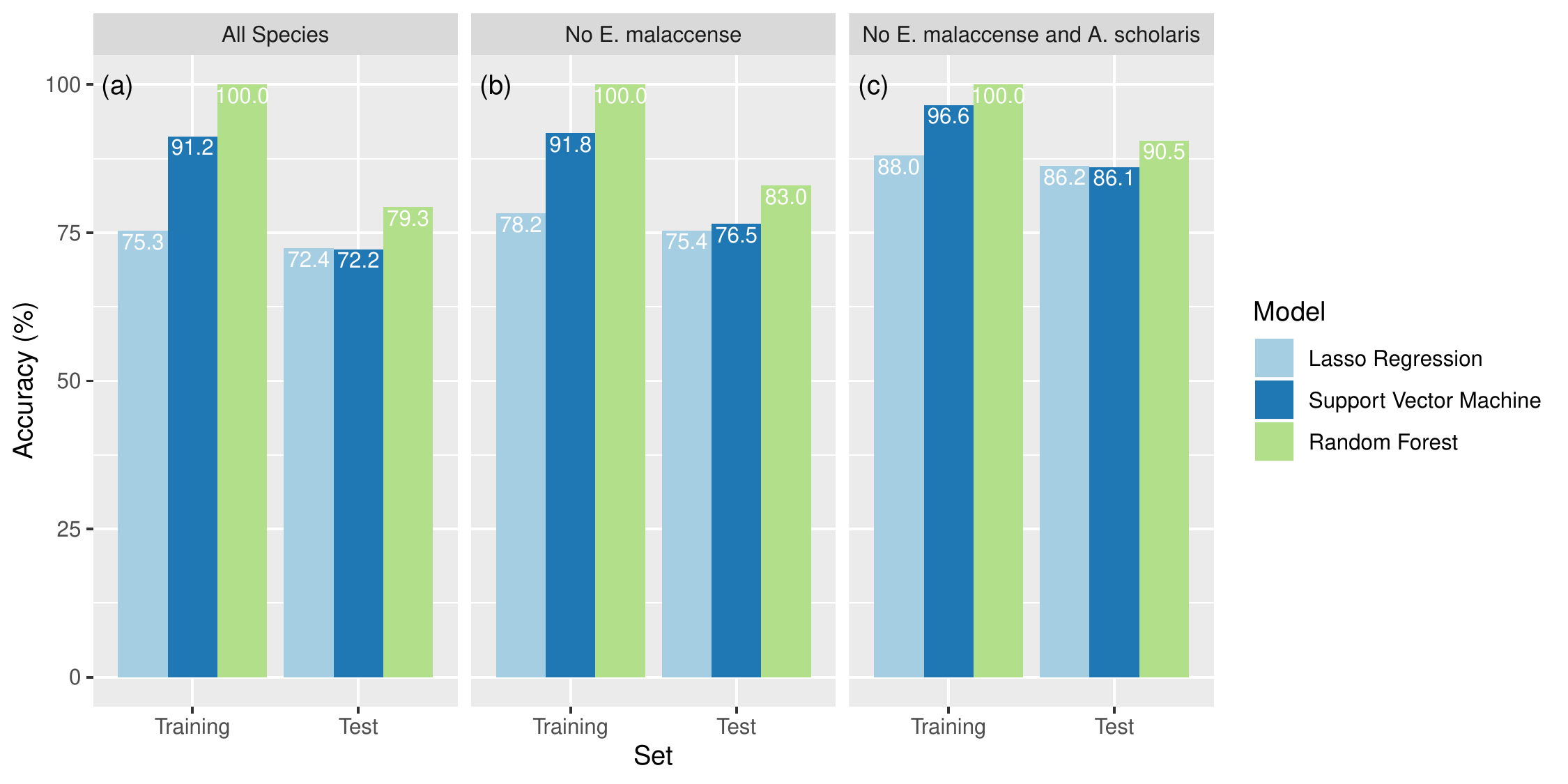}
			\caption{Accuracy of predictions from all three model types (Lasso Regression, Support Vector Machine and Random Forest) on the manually delineated crown dataset. As the number of species was reduced (L to R) the models all improve and the difference between models narrows. Each panel shows both the mean accuracy reported on the crowns used to build the model (training) and on the crowns held back for evaluation (test) in each of the ten cross-validation folds}
			\label{fig:crown_acc}
		\end{center}
	\end{figure}
	
	\begin{table}[h]
		\caption{Prediction accuracies for models based on manually delineated crowns. Accuracy is reported as the mean accuracy on held back data from the ten cross-validation folds, with standard deviation indicated. Best performing model is highlighted in blue.}
		\begin{center}
			\begin{tabular}{lccc}
				& \multicolumn{3}{c}{\textbf{Species labels used}}\\
				\textbf{Model} & All Species & No \textit{E. malaccense} &No \textit{E. malaccense} and \textit{A. scholaris}\\
				\hline
				Lasso Regression & 72.4 $\pm$ 0.4\% & 75.4 $\pm$ 0.4\% & 86.2 $\pm$ 0.3\%\\
				Support Vector Machines & 72.2 $\pm$ 0.7\% & 76.5 $\pm$ 0.4\% & 86.1 $\pm$ 0.4\%\\
				Random Forests & \cellcolor{blue!25}79.3 $\pm$ 0.3\% & \cellcolor{blue!25}83.0 $\pm$ 0.5\% & \cellcolor{blue!25}90.5 $\pm$ 0.3\%\\
				\hline
			\end{tabular}
		\end{center}
		\label{table:res_crown}
	\end{table}
	
	Applied to manually-delineated crowns, the random forests method was most accurate. When assessing the accuracy on the training data there was a hierarchy with random forests models perfectly fitting the training set, while SVM outperformed lasso regression (Figure \ref{fig:crown_acc}, all approaches had accuracy of at least 75\%). When predicting on the held-back test data in each cross-validation fold, random forests modelling performed best on the full seven label model including both long-lived early-successional species (79.3\%), and also on the dataset with \textit{E. malaccense} moved to `other' species (83.0\%) and with \textit{A. scholaris} also moved (90.5\%) (Table \ref{table:res_crown}). Improved accuracy is expected when reducing the number of species labels, but was likely enhanced by the distinctive appearance of \textit{M. gigantea} and \textit{B. pentamera} in contrast to the long-lived early-successional species sequentially moved to `other' species. Generally performance gaps reduced for the test data, suggesting some over-fitting of the more complex models. This is to be expected when working with a dataset of only 633 crowns given the large number of features included, which was highlighted by lasso regression having the smallest drop-off: this methods uses only 25 variables in comparison to 2120 variables afforded to the other modelling approaches. These results showed the overall applicability of the modelling approaches to mapping early-successional species at Harapan, with particular strength in predicting the species indicative of disturbance, with more difficulty on distinguishing long-lived early-successional species from other canopy species. This motivated a particular focus on detecting the indicative species when moving to complete automated mapping, without the need for manual crown delineation.

	\subsection{Automated Species Mapping With SLIC-UAV}
	\label{ssec:res_seg}
	
	\begin{figure}[ht]
		\begin{center}
			\includegraphics[width=\textwidth]{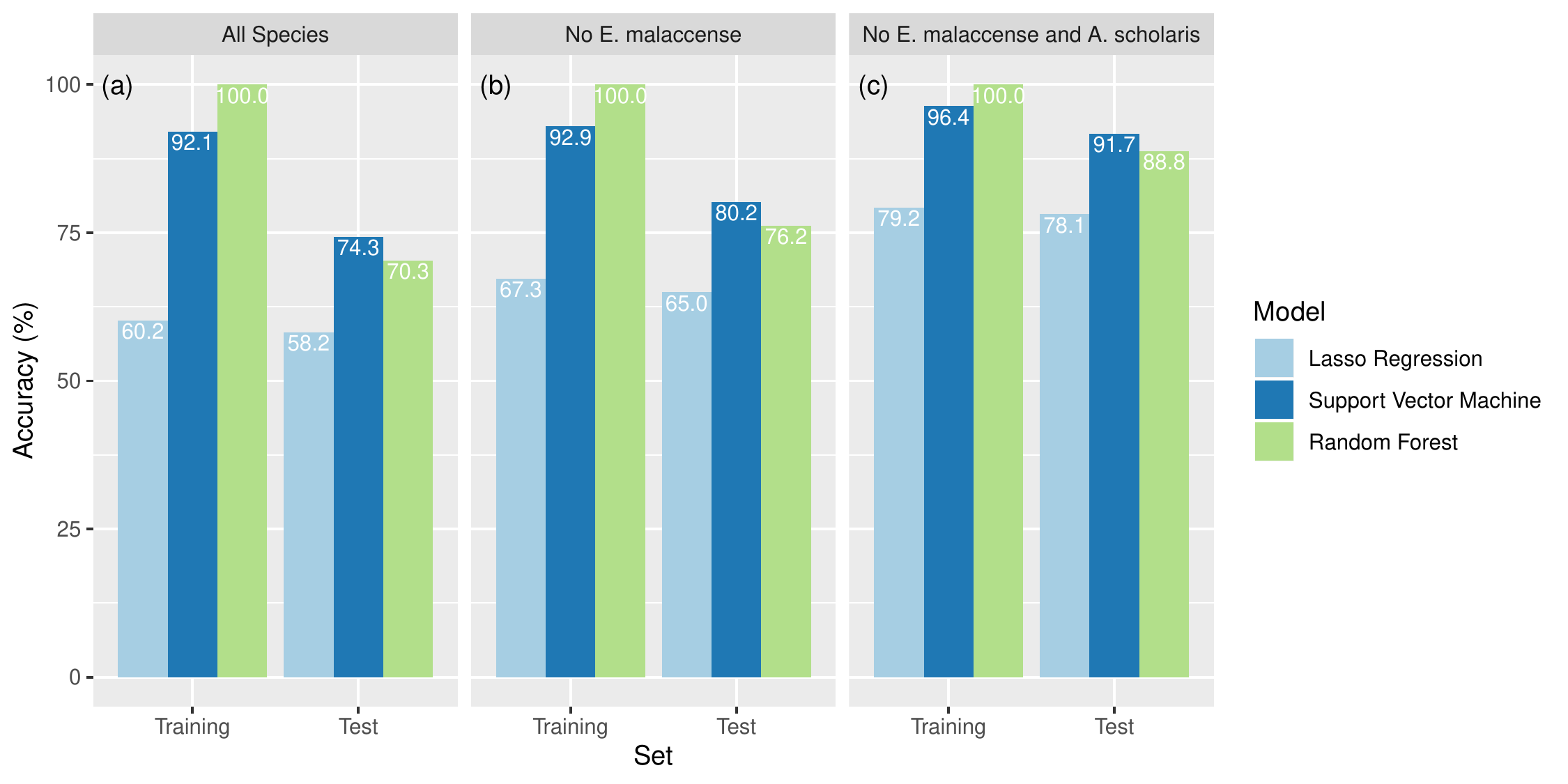}
			\caption{Accuracy of predictions from all three model types (Lasso Regression, Support Vector Machine and Random Forest) on the automatically delineated segments dataset. As the number of species was reduced (L to R) the models all improve and the difference between models narrows. Each panel shows both the mean accuracy reported on the crowns used to build the model (training) and on the crowns held back for evaluation (test) in each of the ten cross-validation folds}
			\label{fig:seg_acc}
		\end{center}
	\end{figure}
	
	\begin{table}[h]
		\caption{Prediction accuracies for models based on automatically delineated segments. Accuracy is reported as the mean on held back data from the ten cross-validation folds, with standard deviation indicated. Best performing model is highlighted in blue.}
		\begin{center}
			\begin{tabular}{lccc}
				& \multicolumn{3}{c}{\textbf{Species labels used}}\\
				\textbf{Model} & All Species & No \textit{E. malaccense} &No \textit{E. malaccense} and \textit{A. scholaris}\\
				\hline
				Lasso Regression & 58.2 $\pm$ 0.8\% & 65.0 $\pm$ 0.6\% & 78.1 $\pm$ 0.4\%\\
				Support Vector Machines & \cellcolor{blue!25}74.3 $\pm$ 0.6\% & \cellcolor{blue!25}80.2 $\pm$ 0.4\% & \cellcolor{blue!25}91.7 $\pm$ 0.3\%\\
				Random Forests &  70.3 $\pm$ 0.8\% & 76.2 $\pm$ 0.3\% & 88.8 $\pm$ 0.2\%\\
				\hline
			\end{tabular}
		\end{center}
		\label{table:res_seg}
	\end{table}
	
\begin{figure}[ht]
		\begin{center}
			\includegraphics[width=0.9\textwidth]{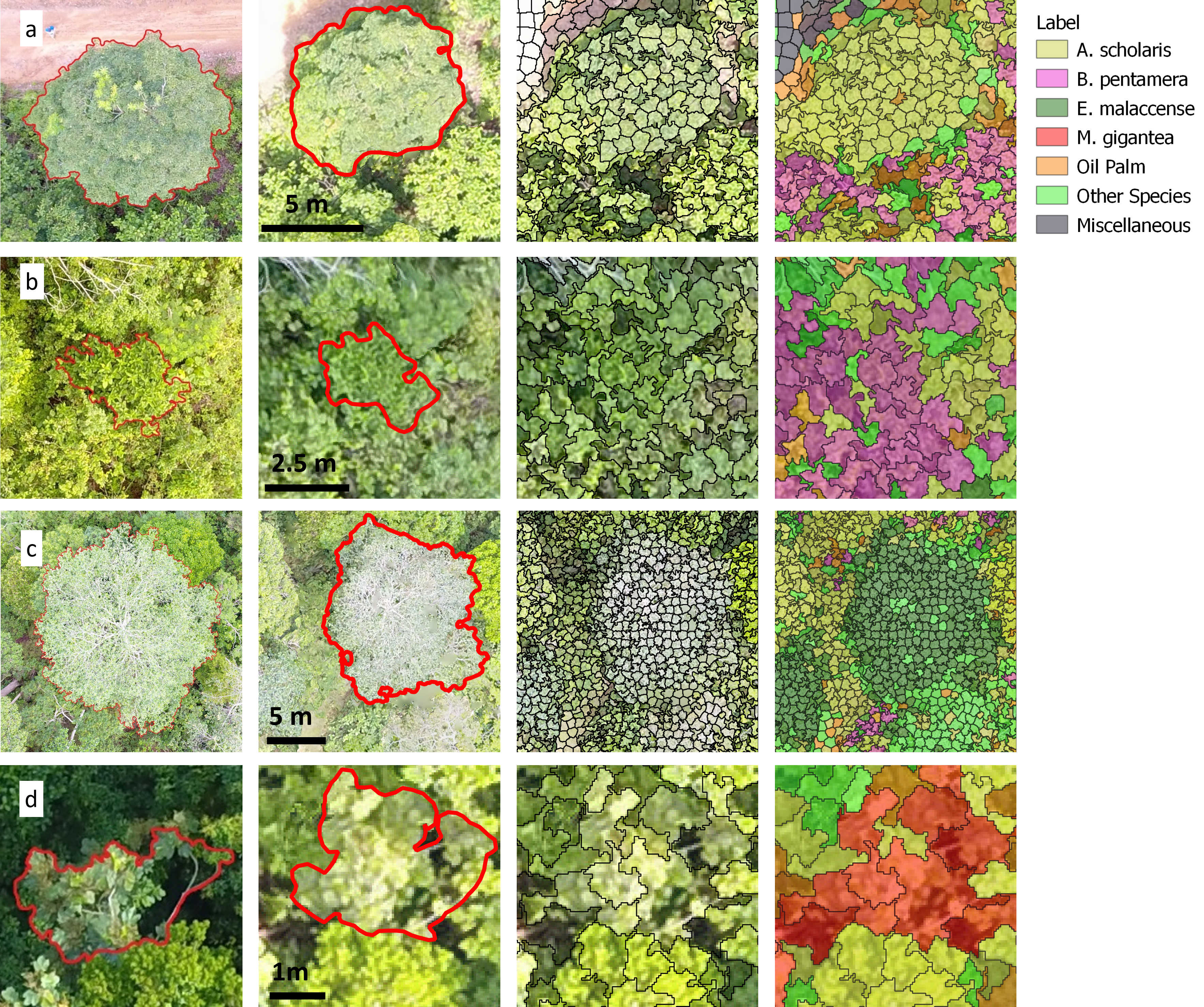}
			\caption{Example images of complete pipeline for the main four species we recorded in the field. Each row includes imagery of one crown: (a) \textit{Alstonia scholaris}, (b) \textit{Bellucia pentamera}, (c) \textit{Endospermum malaccense} and (d) \textit{Macaranga gigantea}. Across the columns (left to right) are shown the manual flight image marked up in the field for crown extent, the outline converted to a shapefile overlain on the RGB orthomosaic, the SLIC superpixels laid over this orthomosaic and the labelling of these superpixels using the SVM approach for the model with all categories. We have only included the outline and labelling for the crown in the centre of each image, but the columns with superpixels are taken from the full landscape map to show this crown in context and so include labels on other superpixels.}
			\label{fig:pipline_grid}
		\end{center}
	\end{figure}

	Once working with the automatically-created superpixels, SVM modelling performed best on all species label options, with random forests outperforming lasso regression. An example of the pipeline for each of the species we mapped in the field is shown in \ref{fig:pipline_grid}. As with crown level modelling, random forests produce 100\% accuracy on training data, SVM comes close to this but lasso regression had a notable reduction in training accuracy compared to crown models (Figure \ref{fig:seg_acc}(a)-(c)). Unlike at the crown-level, the superpixel classifying SVM performed best for all options of labelling (74.3\% accuracy on all species labels rising to 91.7\% for the simplest version with long-lived early-successional species labels removed, Table \ref{table:res_seg}). Random forests modelling performed slightly worse than SVM and lasso regression performed markedly worse. In this case the models with access to all features outperformed the simpler lasso regression in all cases (Table \ref{table:res_seg}). SVM achieved 92\% or higher accuracy on the training data whilst also performing best on test data. In this case the number of labelled regions was much larger than the number of crowns (over 10,000). This enables better use of the algorithms with access to all features, since the number of superpixels is much larger than the number of features, reducing the issues of redundancy of features being used allowing greater over-fitting. Based on this, as well the training scores on the region classification we chose to proceed with SVM as our final model. Notably, the test performance improved markedly for all three modelling approaches when moving to combine all long-lived early-successional species and to focus on early-successional species and oil palm (Figure \ref{fig:seg_acc}(c)). This reaffirms our observation that these species are visually most distinctive, and given their significance as indicators of disturbance, being able to identify these species well is a very valuable advancement.
	
	\clearpage
	
	\subsection{Contribution of Imagery and Features}
	\label{ssec:res_feats}
	
	We investigated the effect on model performance for using different combinations of imagery input and features computed on model performance for the fully-automated SLIC-UAV pipeline. For this we completed a full multiplex of all valid combinations of imagery (RGB, multispectral and DSM) and all valid sets of feature types (spectral and textural). Owing to issues with the multispectral sensor recording illumination, leading to artefacts in the imagery produced (Figure \ref{fig:land_map}(c)), we masked out all crowns which were within a 50 m radius of any issue, marked by the orange hashing, leaving 409 crowns for evaluation. To reduce the workload to build so many models, we chose to use a simpler 75\%/25\% split of the data into training and test sets, using the same for all models. We chose to focus on a sequential addition of imagery in line with the additional processing or sensors that are required to see if these steps are justified by performance.
	
	\begin{figure}[ht]
		\begin{center}
			\includegraphics[width=\textwidth]{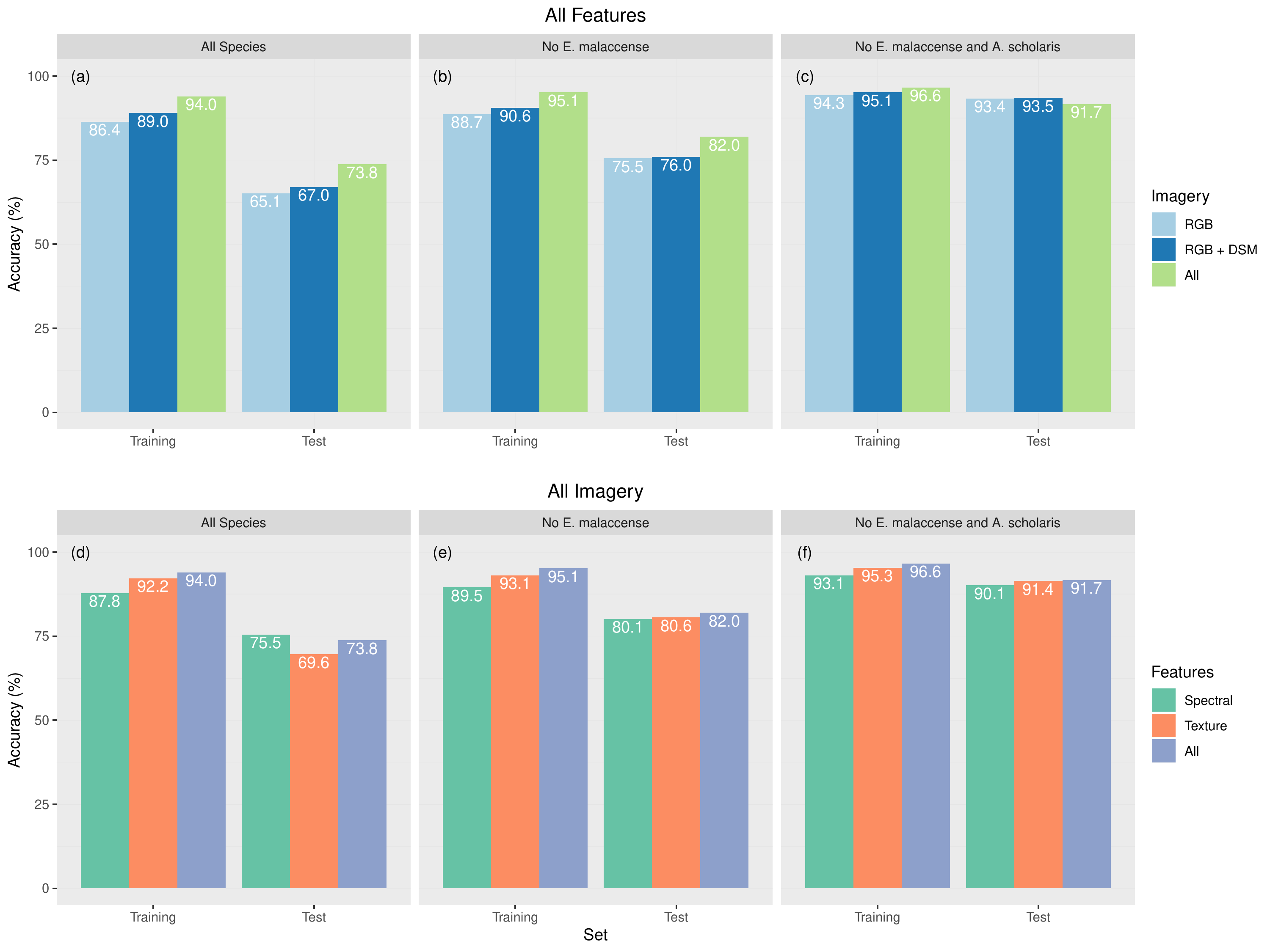}
			\caption{Accuracy of predictions of SVM on superpixels for the three model forms, with species removed going from left to right, using different combinations of imagery and features. The top row (a)-(c) shows the effect of including successively more imagery forms, adding DSM and then MS data to RGB data (using both spectral and textural features). The bottom row (d)-(f) shows how using spectral, textural or both sets of features changes performance (using all three forms of imagery). Groups of bars represent accuracies on the superpixels used for model creation (training) and on the superpixels held back for evaluation (test). Generally inclusion of DSM data improves performance over RGB data alone and multispectral improves model performance in all but the simplest set of species labels. Generally textural features do better than spectral features, except in the model for all species, but combining both of these improves accuracy.}
			\label{fig:seg_feat_imagery}
		\end{center}
	\end{figure}
	
	Changing the imagery used to develop models using all features produced a mostly expected result. For training models the picture was simple and as expected, adding structural and multispectral data sequentially improved accuracy (Figure \ref{fig:seg_feat_imagery}(a)-(c)). When tested on the held back data adding DSM data improved accuracy for all three species label sets. This was expected as these data add more information on vertical structure, whereas MS and RGB data contain information on spectral response and two-dimensional structure, from image contrasts. The addition of multispectral imagery further improved model performance in the two more complex models (Figure \ref{fig:seg_feat_imagery}(a)-(b)) but actually decreased performance on held-back data in the simplest model. This improvement added more to the model than adding DSM data, showing the additional value of these data. Comparing RGB and multispectral data directly suggested a similar pattern (Figure \ref{fig:panel_subsets_imagery}), where comparable models using multispectral imagery in place of RGB performed better whenever textural features were included, except in the case of the simplest problem with only five species labels, as in Figure \ref{fig:seg_feat_imagery}(c). However, when using only spectral features, multispectral imagery over RGB gave better fit on data used to train at the cost of worse fit on held-back data. This suggests the extra information in the multispectral imagery may have led to overfitting when only the pixel values were considered and not their textural context.
	
	Breaking superpixel classification with SVM down by which features were used gave a mixed picture of which variables contribute the most to model accuracy (Figure \ref{fig:seg_feat_imagery}(d)-(f)). Generally including both classes of features did best for all models barring the one with most labels. For the two models with a reduced set of labels, the hierarchy was clear, with textural features producing a slight improvement over spectral features, with the combination doing best (Figure \ref{fig:seg_feat_imagery}(e)-(f)). This wasn\textsc{\char13}t the case for the model with all species labels. Here the best performance occurred for spectral features alone, with textural features doing worst. Adding spectral features to these improved performance, but this was still worse than using spectral features alone.

	\subsection{Landscape Species Mapping with SLIC-UAV}
	\label{ssec:res_land}
	
	Taking forward SLIC-UAV combined with SVM modelling, we built a final model using the labelled crowns to extend our labelled crown examples to a full landscape map. We chose to keep all species labels for a more informative map, and present the confusion matrix of the final model used for landscape mapping, assessed using the mapped crowns, in Table \ref{table:res_seg_mat}. In particular, the model does best in identifying Non-vegetation and oil palm, with both having precision and recall greater than 95\%. Amongst the species we mapped, the earl-successional species indicative of disturbance are identified with highest precision (\textit{Bellucia pentamera}: 90.1\%, \textit{Macaranga gigantea}: 96.9\%) whereas the longer-lived early-successional species are more commonly confused with each other as well as the `other' species category, leading to lower precision (\textit{Alstonia scholaris}: 86.4\%, \textit{Endospermum malaccense}: 89.5\% and `others': 87.5\%). Recall values were generally lower, in most cases owing to mis-classification of known species as the `other' category, being a catch-all category this is unsurprising. The value of including this label enables a `control' label for vegetation, ensuring that superpixels identified as a species of interest are done so with high precision, since the model is not forced to choose between a limited set of species labels. This high precision is of value to management as it gives high confidence that regions predicted as including species of particular concern do indeed feature this species. 
	
\begin{figure}[h]
		\begin{center}
			\includegraphics[width=\textwidth]{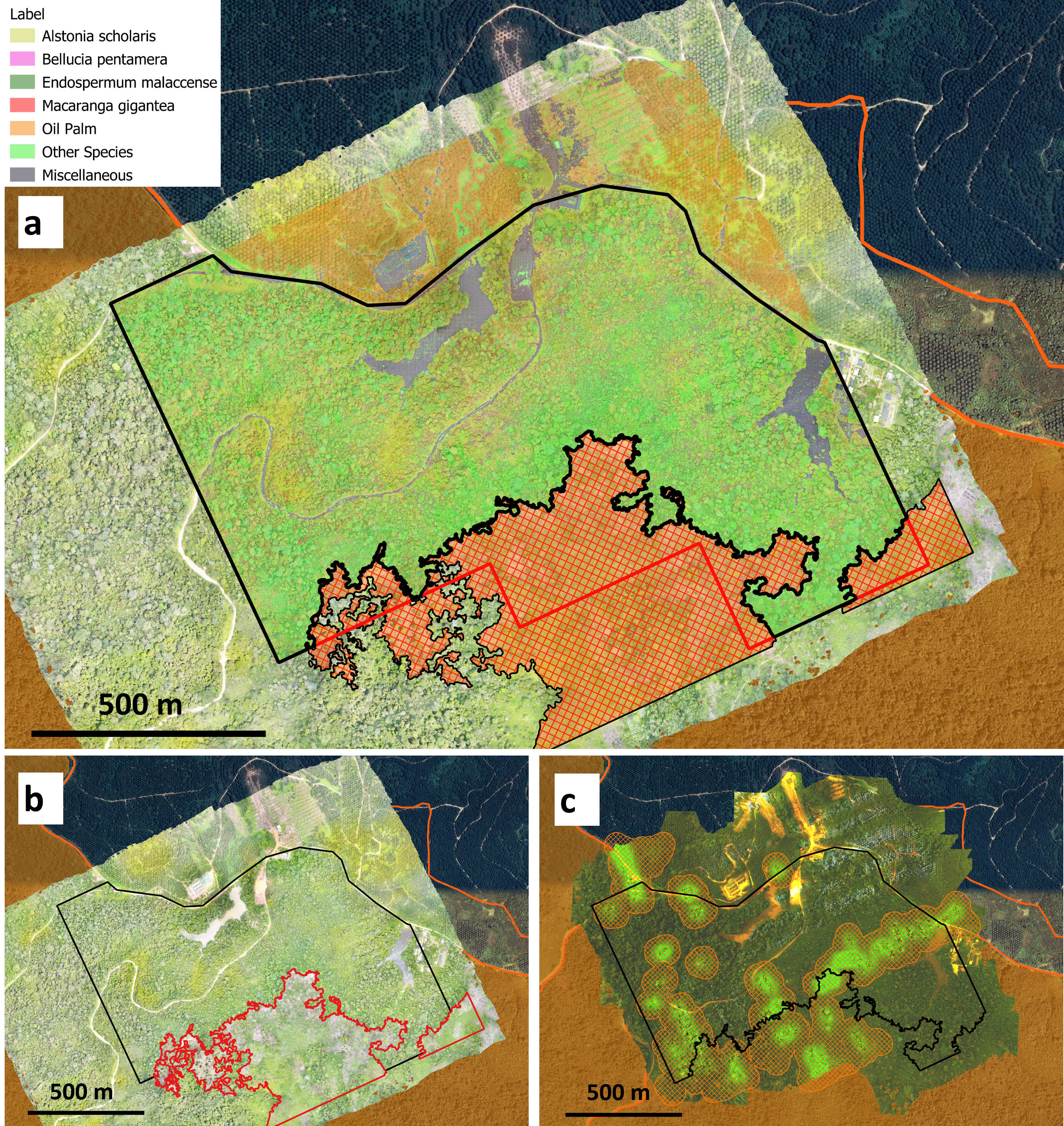}
			\caption{Map showing the predictions form the full seven labels model using SLIC-UAV and Support Vector Machines across the region used in this work. Panel (a) shows the map of species labels, where the distinction of plantation (mostly oil palm), non-vegetated and forest is clearly picked up, with \textit{A. scholaris} accounting for a large amount of the forest region, especially close to roads, as is common for this species. Black outline shows area within Hutan Harapan used for Table \ref{table:res_land}. The red crossed-out region represents the region we masked owing to forest clearing between the two field seasons, with the red line across this indicating the area where imagery was collected in both surveys. Panel (b) shows the RGB imagery used in this study, with the masked out region again in red. Panel (c) shows the multispectral imagery as a false colour (Red as red, Green as green and Red-edge as blue). The artefacts from processing the correction of this imagery can be seen by bright areas. The buffered region where we excluded regions when multiplexing features is shown as orange crosshatching.}
			\label{fig:land_map}
		\end{center}
	\end{figure}

\begin{figure}[h]
		\begin{center}
			\includegraphics[width=\textwidth]{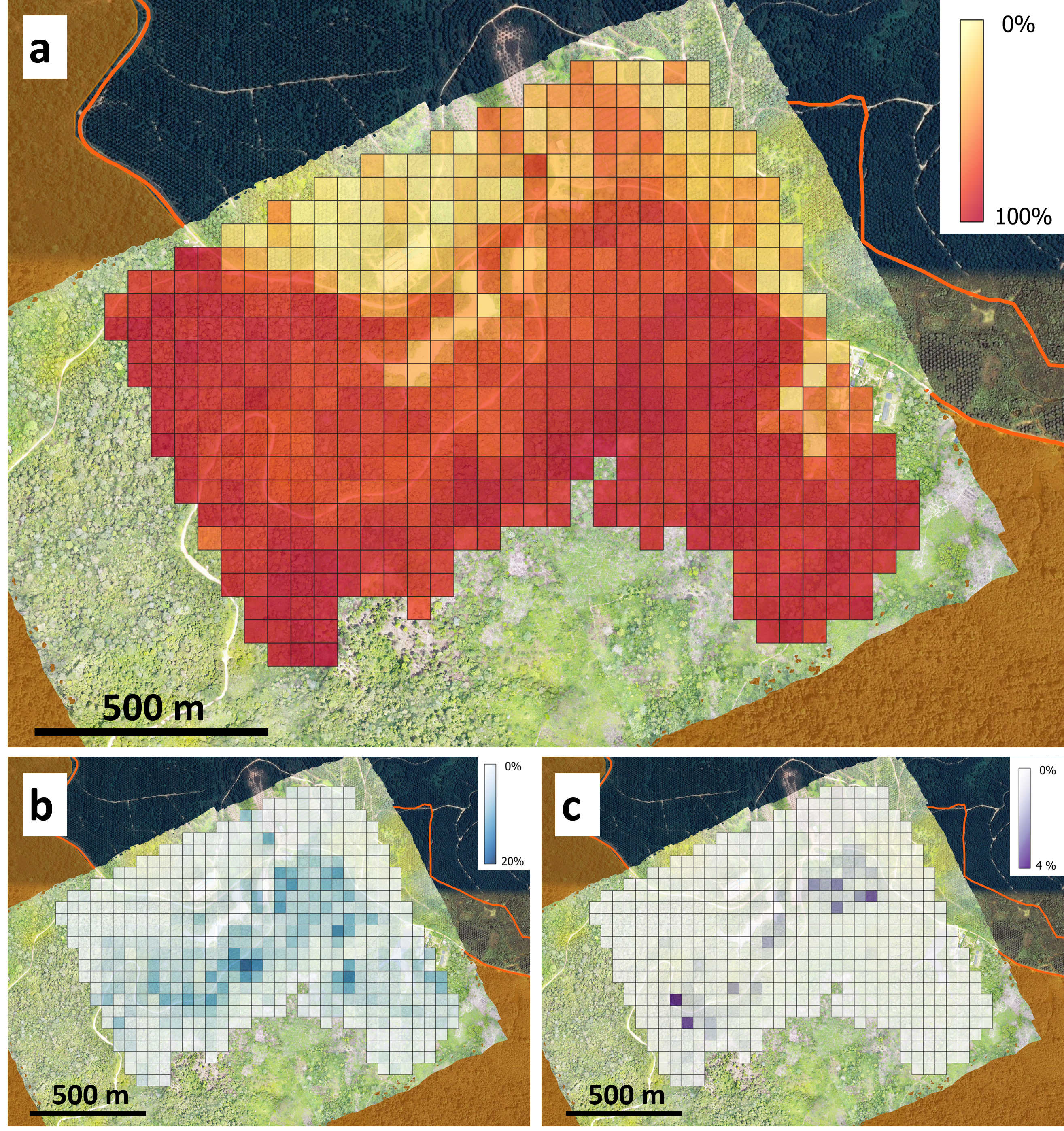}
			\caption{Map of the percentage occurrence by area of vegetation species in 0.25 ha cells across the study area as predicted by the superpixel SVM model including all species (as used for Figure \ref{fig:land_map}). Panel (a) shows the combined occurrence for long-lived early-successional species for \textit{Alstonia scholaris}, \textit{Endospermum malaccense} and `other' tree species, with panel (b) showing \textit{Bellucia pentamera} alone and panel (c) showing \textit{Macaranga gigantea} alone.}
			\label{fig:land_map_BEL_MAC}
		\end{center}
	\end{figure}

	The map of predictions from our most complete SVM model applied to superpixels (with all species) over the full landscape is shown in Figure \ref{fig:land_map} along with the input RGB and multispectral imagery used, in which the region with clearing between the surveys is masked out. The broad classification pattern corresponds precisely with the site history, with oil palm, non-vegetation and forest all well-identified. Notably, the oil palm plantation at the top of the region, lying outside the Hutan Harapan boundary, shows the specificity of our model, with regions of bare ground, buildings and remaining vegetation clearly picked up in contrast to the oil palm. After clipping the map to just the area within Hutan Harapan, the early-successional species together comprised 45.34\% of the study site with oil palm and other vegetation represented 7.21\% and 41.81\% of the site respectively (Table \ref{table:res_land}). \textit{A. Scholaris} represented 38.37\% of the study area most commonly occurring close to roads. The other species of interest made up 6.97\% of the cover, or 8.00\% of the cover once non-vegetation and oil palm are removed. The remaining forest species covered the remaining 41.81\%. Higher occurrence of \textit{B. pentamera} and \textit{M. gigantea} was expected, but these species tend to dominate the sub-canopy and appear more rarely as large `top of canopy' trees. These were often identified in canopy gaps and near the edge of crowns. Gridded maps of canopy dominance for the combination of long-lived early-successional species with `other' vegetation, \textit{Bellucia pentamera} and \textit{Macaranga gigantea} (Figure \ref{fig:land_map_BEL_MAC}) (a)-(c), reveal increased occurrence of these early-successional species close to roads, where disturbance is highest. Generally, there was also a slight gradient for increased prevalence of long-lived early-successional species further from the boundary of Hutan Harapan, owing in part to reduced sporadic occurrences of oil palm moving further from the plantation (Figure \ref{fig:land_map_BEL_MAC}).
	
	\begin{table}[h]
		\caption{Confusion matrix for Support Vector Machine model applied to the automatically delineated SLIC-UAV superpixels using all labels, as used for landscape mapping.}
		\begin{center}
			\resizebox{\columnwidth}{!}{%
				\begin{tabular}{c|c|>{\centering\arraybackslash}m{1.5cm}>{\centering\arraybackslash}m{1.7cm}>{\centering\arraybackslash}m{1.9cm}>{\centering\arraybackslash}m{1.5cm}>{\centering\arraybackslash}m{1.5cm}>{\centering\arraybackslash}m{1.5cm}>{\centering\arraybackslash}m{2cm}|c}
					\multicolumn{2}{c}{} & \multicolumn{7}{c}{\textbf{Actual label}} & \\
					\hhline{~~-------~}
					\multicolumn{1}{c}{} & & \textit{Alstonia scholaris} & \textit{Bellucia pentamera} & \textit{Endospermum malaccense} & \textit{Macaranga gigantea} & Oil Palm & Non-vegetation & Other \hspace{1cm} Vegetation & \textbf{Precision} \\
					\hhline{~--------~}
					& \textit{Alstonia scholaris} & 2288 & 57 & 149 & 16 & 1 & 8 & 130 & 86.4\% \\
					& \textit{Bellucia pentamera} & 26 & 617 & 16 & 9 & 0 & 5 & 12 & 90.1\% \\
					\textbf{Predicted}& \textit{Endospermum malaccense} & 36 & 18 & 1026 & 0 & 1 & 0 & 66 & 89.5\% \\
					\textbf{label}& \textit{Macaranga gigantea} & 1 & 2 & 1 & 218 & 0 & 1 & 2 & 96.9\% \\
					& Oil Palm & 3 & 3 & 0 & 5 & 2280 & 6 & 7 & 99.0\% \\
					& Non-vegetation & 0 & 0 & 0 & 0 & 0 & 1893 & 0 & 100.0\% \\
					& Other Vegetation & 119 & 45 & 121 & 38 & 4 & 61 & 2705 & 87.5\% \\
					\hhline{~--------~}
					\multicolumn{1}{c}{} & \multicolumn{1}{c}{\textbf{Recall}} & 92.5\% & 83.2\% & 78.1\% & 76.2\% & 99.7\% & 95.9\% & \multicolumn{1}{c}{92.6\%} &  \\
				\end{tabular}
			}
		\end{center}
		\label{table:res_seg_mat}
	\end{table}
	
	\begin{table}[h]
		\caption{Abundance of species predicted across the Harapan landscape, with the oil palm plantation in the North and regions of forest clearance removed as shown in Figure \ref{fig:land_map}}
		\begin{center}
			\begin{tabular}{lcc}
				\hline
				Class (Full Model) & Cover (\%) & Cover (ha) \\
				\hline
				\textit{Alstonia scholaris} & 38.37 & 35.96\\
				\textit{Bellucia pentamera} & 4.26 & 3.99\\
				\textit{Endospermum malaccense} & 2.42 & 2.26\\
				\textit{Macaranga gigantea} & 0.29 & 0.27\\
				Oil Palm & 7.21 & 6.75\\
				Non-vegetation & 5.64 & 5.29\\
				Other Vegetation & 41.81 & 39.18\\
				\hline
				TOTAL & 100.00 & 93.70\\
				\hline
			\end{tabular}
		\end{center}
		\label{table:res_land}
	\end{table}
	
	\clearpage


	\section{Discussion}
	\label{sec:dis}
	
	We were able to detect early-successional vegetation in recovering logged-over tropical forests with accuracies of 79.3\% and 74.3\% for hand-drawn crowns and automatically segmented superpixels respectively. Focusing only on early-successional species with visually-distinctive features boosted accuracy to 90.5\% and 91.7\% respectively. This suggests that while identifying visually-distinct early-successional species is highly tractable, the similarity of long-lived early-successional species to more typical forest species, in terms of their spectral reflectance and canopy architecture, makes identifying them more challenging. Together this highlights the value of assessing biodiversity, during forest restoration, through the abundance and composition of early-successional species but the efficacy of the approach depends on the extent to which indicator species are distinct from the background of more typical forest species.
	
	\subsection{Crown-level Species Models}
	\label{ssec:dis_crown}
	
	Working with manually digitised crowns from species of management interest, we found our choice of spectral and textural features led to models with performance competitive with the cutting edge in this established problem of species identification for known crowns in the challenging and diverse environment of recovering tropical forest. By using random forests with our features we achieves cross-validation accuracies of 79.3\%, 83.0\% and 90.5\% for models with six, five and four categories for vegetation respectively (Table \ref{table:res_crown}). The improvement came from merging species labels for the long-lived early-successional species \textit{Alstonia scholaris} and \textit{Endospermum malaccense} into the general vegetation category, with the large jump for the simplest model suggesting particular success with discriminating the earl-successional species indicative of disturbance in tropical secondary forest which are of particular interest to management of restoration projects. These results are competitive with other studies on identifying mapped crowns from UAV imagery: \citep{Lisein2015} had accuracies ranging from 64\% to 84.1\% for identifying five species for ten independent surveys of RGB and multispectral data of the same trees, \citep{Michez2016} identified four species and a general vegetation label with 60.2\% and 48.5\% accuracy for two strips of riparian vegetation from UAV multispectral imagery whereas \citep{Alonzo2018} mapped four species with an additional shrub category in coniferous forests using only RGB imagery with 85\% accuracy, though this required additional data from above ground canopy height, which requires canopy gaps to be robustly measured, which are rare in logged-over tropical forests, making accurate estimation difficult, for example see \citet{Swinfield2019} for this problem within Hutan Harapan. Combining this with the increased complexity of secondary tropical forest, we believe our method remains competitive. In a more controlled setting \citep{Tuominen2018} identified crowns of 26 species in a Finnish arboretum, though most had fewer than 20 examples, all grown in homogeneous stands. Using RGB imagery and structural data from photogrammetry, overall accuracy was 63.7\%, but rose to 68.3\% when illumination was corrected. In this same study, the inclusion of hyperspectral data on the UAV improved accuracy to 74.9\% before correction and 78.3\% after, highlighting the value added by these data and stressing the case for development of cheaper sensors. Clearly there is considerable diversity in the sensors, analytical pipelines and forest ecosystems being tested, which makes direct comparison challenging, even when just considering hand-drawn crowns. This validation removes the complexity of superpixel segmentation, where imperfect boundaries can introduce noise and assesses the power of the features we derive to build accurate models for species mapping. Yet, this task is only a stepping stone towards the wall-to-wall mapping approaches needed for evaluating biodiversity during forest restoration, since manually digitising crowns is a labour intensive process and does not scale well to management units and landscapes.

	\subsection{Automated Species Mapping With SLIC-UAV}
	\label{ssec:dis_seg}
	
	The challenge in tracking biodiversity recovery during forest restoration is producing species classification methods that generalise over large contiguous areas without need for exhaustive surveys. Using our new SLIC-UAV pipeline combined with predictive modelling, as explored in \ref{ssec:dis_crown}, we were able to identify our five species of interest from among other vegetation and non-vegetated regions with 74.3\% accuracy, rising to 91.7\% when we focused only on species indicative of recent disturbance: \textit{Bellucia pentamera} and \textit{Macaranga gigantea} (Table \ref{table:res_seg}). The power of this approach is that it can then directly be applied to superpixels without an existing species label, enabling extension of labelled data across a management unit of 100 ha. This requires a method which can reliably be extended to unseen data, so validation within the labelled data was crucial to our final output of maps for key species indicative of disturbance and recovery trajectories. Support vector machines (SVM) performed best (Figure \ref{fig:seg_acc}) and increased in accuracy compared with manually-delineated crown modelling with lower overfitting, showing stronger robustness to variation introduced by working with superpixels. In contrast lasso regression and random forests showed a loss in accuracy, with random forests having more problems of overfitting that in the simpler defined crown problem. Splitting of imagery by SLIC follows contrast boundaries, which naturally occur within crowns owing to different components (leaves and branches) as well as variation within these owing to factors like health and illumination. This leads to superpixels with local homogeneity where differences within a crown are split causing dilution of the boundaries between each species. There are more labelled regions, with more apparent variation within species, making species discrimination more complex than for whole crowns. Correct labelling of superpixels requires adaptability to this structure, without risk of overfitting species boundaries. We believe this is why SVM did best, by using a non-linear kernel for boundaries whilst regularising to control overfitting. In our study, lasso regression was not flexible enough to adapt to this more complex problem, and random forests, being less regularised, struggled with the intra-species variability of superpixels and overfitted. This is also consistent with our observations that the early-successional species have much more distinct textures than \text{Alstonia scholaris} and \textit{Endospermum malaccense}. Thus, the boundaries between these two species and the `other species' category are the least clear in the updated problem, leading to more obvious overfit by all three approaches on unseen data (Figure \ref{fig:seg_acc}(a)-(b) compared to the simplest model \ref{fig:seg_acc}(c). This justifies our use of SVM for subsequent landscape mapping, balancing accuracy with robustness to the need to extend the model beyond the data it is trained on to ensure output maps of forest status are reliable and of use to project managers.
	
	Before moving forward to see the contribution SLIC-UAV with SVM modelling can bring to assessing secondary tropical forest recovery we first consider the accuracy of our approach in comparisons with similar pipelines published in other studies for species occurrence mapping from UAV data. Modelling approaches include: CART decision trees \citep{Feduck2018,Wu2019}, k-nearest neighbours \citep{Tuominen2018,Wu2019}, neural networks \citep{Colkesen2018,Fujimoto2019,Kattenborn2019a,Kattenborn2019b,Osco2020}, random forests \citep{Feng2015,Lisein2015,Michez2016,Goodbody2018a,Colkesen2018,Roder2018,Tuominen2018}, regression variants \citep{Alonzo2018}, support vector machines \citep{Colkesen2018,Wu2019}, although results are setting dependent and there is no clear best modelling method. We focus on analysis pipelines most closely related to our work. \citet{Michez2016} combined segmentation using eCognition software and spectral and textural features of multispectral imagery to identify four species of interest in riparian strips with random forests with 79.5\% and 84.1\% for their best model at each of their two sites. \citep{Franklin2018} also mapped four deciduous species in Canada using multispectral imagery with ENVI software to do automatic region extraction, requiring manual adjustment and merging to match labelled crowns. Using random forests on 109 total crowns, they achieved a test accuracy of 78\%. Both of these problems map most closely to our problem with all labels excluding \textit{Endospermum malaccense}, where we achieved 80.2\% accuracy for a full automated pipeline, in a complex secondary tropical forest. This makes our method competitive on the fully supervised problem of correctly identifying species for imagery objects. However, our pipeline is built to map directly to semi-supervised mapping, where only a small fraction of a whole area is manually identified without requiring any manual adjustment or selection of objects and superpixels as we show in our landscape mapping in \ref{ssec:res_land}. The real power of our method is ability to extend labelled crowns to map species abundance across a block of forest, 100 ha in our case; something not addressed in the works we have compared to, but crucial for these methods to aid management of restoration effort. Following this, a key strength of our pipeline is identifying early-successional species indicative of recent disturbance: the pioneer \textit{Macaranga gigantea} and invasive \textit{Bellucia pentamera}. This success is reflected in improved accuracy of 91.7\% when grouping all long-lived early-successional species (Figure \ref{fig:seg_acc}(c)), and the high precision in identification of these species when moving to landscape mapping (Table \ref{table:res_seg_mat}). This is a simpler problem of focusing only on three species of interest, the two indicative species and oil palm, for which there are more comparable methods, looking only at one or two species of interest from UAV imagery. \citet{Wu2019} mapped an invasive species on a Chinese island using eCognition to generate objects from UAV RGB imagery, validated on a per-pixel basis with 95.6\% overall accuracy and 94.4\% precision for the invasive species, though this was from focus on only one species. \citep{Apostol2020} also used OBIA through eCognition to identify regions as either spruce or birch in Romanian forests with accuracy varying from 73.9\% to 77.3\%. Clearly accuracy can be very context dependant, but we feel that in the complex environment of logged-over recovering tropical forest, our accuracy of 91.7\% compares favourably. 
	
	SLIC-UAV had success on detecting and mapping early-successional species indicative of disturbance, but reliability of detection of long-lived early-successional species was hindered by the visual similarities of these species. This finer difference may be better picked up by more bespoke features as opposed to our use of features based on existing approaches in image analysis. An alternative approach which we believe may help with expanding available features leading to better discrimination of these species are convolutional neural networks. These are able to create complex features fully automatically which may help discriminate the more visually similar species. As a trade off, they approach normally require more computational power to develop and fit models, and are constrained by the need for a regular input shape. Recent works show the power of such approaches: \citet{Fujimoto2019} extracted cedar and cypress crowns in a Japanese forest with an automated method, before using neural networks to classify standardised images using only the equivalent of our DSM imagery with an accuracy of 83.6\%, much higher than any approach using only DSM imagery in Figure \ref{fig:panel_subsets_imagery} and \citet{Kattenborn2019a} used neural networks to map two species form RGB UAV imagery across a successional gradient in Chile, with accuracies of 87\% and 84\% and extended this work in \citet{Kattenborn2019b} to instead map percentage species occurrence in cells of a grid, which may be a more computationally feasible way to build maps to guide management in such an approach. These accuracies compare favourably for SLIC-UAV, but show the power of a method with no explicitly computed features. A future step for SLIC-UAV could be to work with training a neural network to distinguish differences between the most hard to distinguish species and to use this to help construct or learn features for modelling to help this distinction. We also didn't consider the predicted species for neighbouring superpixels when predicting each superpixel, but we would expect a strong local correlation, given few crowns are as small as our superpixels. A processing step which then allows adjustment based on confidence of prediction relative to local superpixels and their predicted species label could improve robustness of the predictions. This approach has been shown to improve accuracy by as much as 4\% \citep{Tong2019}, but we haven\textsc{\char13}t tested this here
	
	\subsection{Contribution of Imagery and Features}
	\label{ssec:dis_feats}
	
\begin{figure}[ht]
		\begin{center}
			\includegraphics[width=\textwidth]{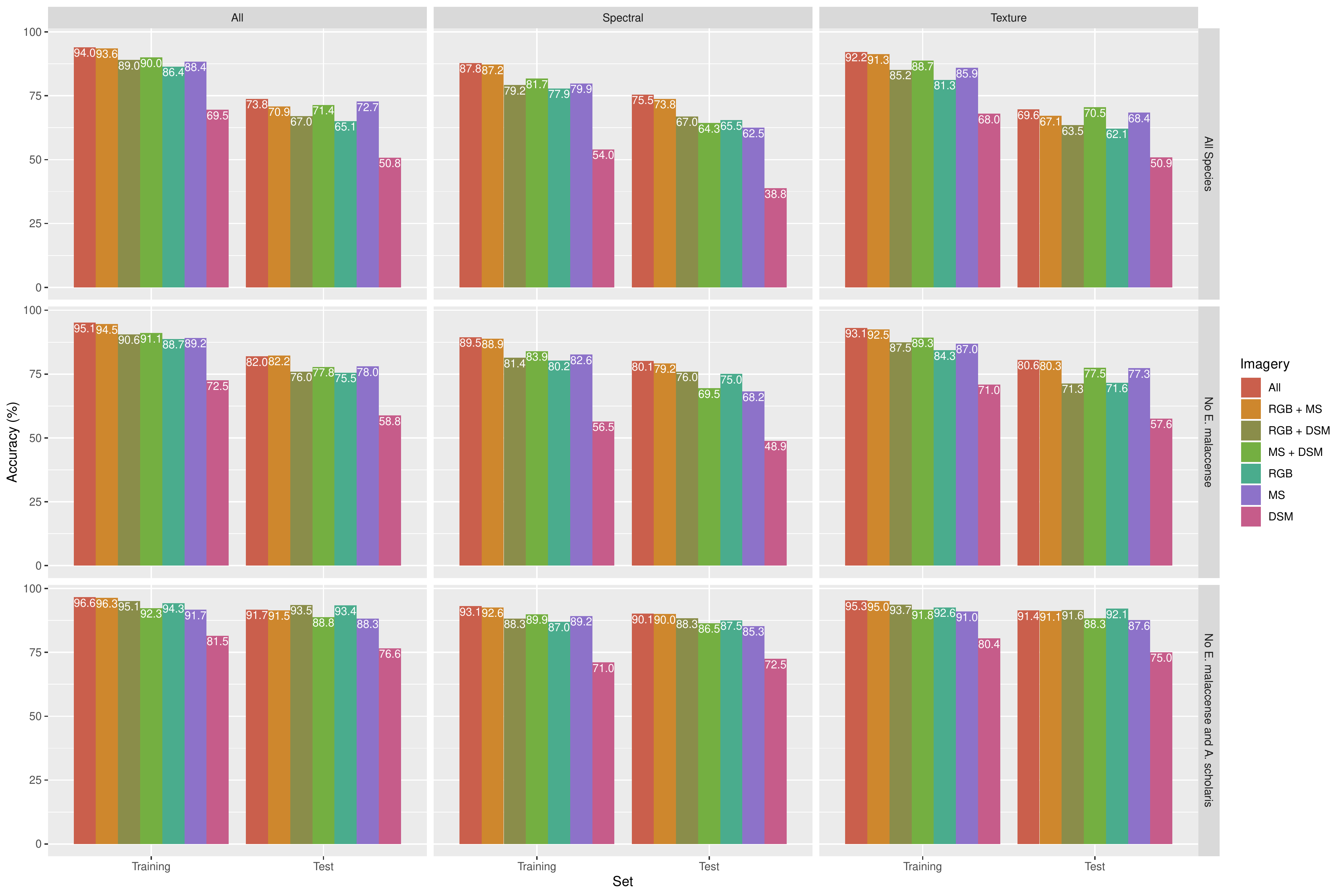}
			\caption{Accuracy of predictions from SVM on training and test SLIC-UAV superpixels data for the three model forms, using different combinations of imagery and features. Here, only superpixels from crowns unaffected by multispectral processing artefacts were used, and a simple 75\%:25\% training to test split was used, being the same in all models. Lattice rows represent the model forms, lattice columns represent feature classes used, colour of bars represents the imagery used and groups of bars represent training and test accuracies. As before, simpler models generally do better. Removing Imagery sources reduces accuracy with DSM data alone giving weakest performance. Multispectral imagery generally adds more to performance than comparable models using RGB imagery, except in the problem with fewest labels (bottom row) and where only spectral features are used.}
			\label{fig:panel_subsets_imagery}
		\end{center}
	\end{figure}

	We found that progressively adding more forms of imagery generally improved the accuracy of resulting models (Figures \ref{fig:seg_feat_imagery} and \ref{fig:panel_subsets_imagery}). This is unsurprising, as the addition of DSM imagery improves the detail in structural data, including the third dimension of data, and adding multispectral imagery added two additional regions of spectral response, namely red edge and near infrared, being regions of the spectrum for which vegetation reflect a greater proportion of incident light, supporting findings in existing works such as \citet{Michez2016} and \citet{Lisein2015} who found models worked best when including both RGB and multispectral data. The exceptions to these broad trends were models focusing only on the simplest problem, removing the distinction of \textit{Alstonia scholaris} and \textit{Endospermum malaccense} from `other' species. These saw an improvement in fit for each imagery source added in fitting the training data, but the addition of multispectral imagery led to poorer performance on held back data (figure \ref{fig:seg_feat_imagery}(c)). We believe this may owing to the greater variation in the multispectral data. This particular model already achieves performance above 90\% without addition of these data, and the improvement in training accuracy is much smaller than for the models with more species categories. It may be that this very minimal gain in accuracy on training data comes with a trade off of reduced generalisability, owing to sensitivity to the bigger differences between species in the red edge and near infrared spectral range. This possible mechanism is supported by our full multiplex analysis in Figure \ref{fig:panel_subsets_imagery}. Here the models which only use features base don the spectral properties of each superpixel all show better fit on the training data for models using multispectral imagery compared to RGB, but universally worse performance on held back superpixels. this suggests models using only the spectral response from multispectral imagery may be prone to overfitting, so it is important to consider the quality of these data and how they are used. Here we have already processed the multispectral imagery to correct for illumination, without with performance is reduced, such as in \citet{Tuominen2018}. Addition of more advanced sensors, such as for hyperspectral imagery, may further improve results, but are often prohibitively expensive for this context or custom made \citep{Hruska2012,Colomina2014,Aasen2015}. We conclude that this imagery can add to performance in species mapping models, but must be considered carefully, and certainly a consideration of the benefits in accuracy compared to the additional cost for such sensors is worth making for any application of our work.
	
	It is also important to consider the ways that imagery is processed for each superpixel to create feature scores. As shown in Figure \ref{fig:seg_feat_imagery}, textural features based on patterns in each superpixel generally allow better discrimination on training examples than using only the statistics of spectral responses, but that combining these is best. This pattern is consistent for our two simpler models on held-back data. This can be easily reasoned. For these models the early-successional species, with distinctive textures, are increasingly important and so texture is key. However, adding spectral information can still help discrimination. This is not the case for our model with the most species labels, where spectral data alone perform best, and adding textural measures actually reduces performance. This is surprising, suggesting the addition of textural features to a spectral model alone makes the classification boundaries less clear. We believe this follows from confusion with the species \textit{Endospermum malaccense} and the early-successional species. The pattern of leaves for many crowns of this species have a very `jagged' appearance, very similar to that of \textit{Bellucia pentamera}. The spectral signature of the two are more distinctive, but we believe, using subsets of \textit{E. malaccense} crowns as superpixels can appear texturally similar to \textit{B. pentamera}. This problem disappears once the label of \textit{E. malaccense} is removed, leading to a different global classification boundary landscape. With more examples of \textit{E. malaccense} we would hope this boundary would be better defined, as the increased weighting given to this label to balance training examples may contribute to this confusion. Overall we feel results justify inclusion of textural information, with the noted confusion in one case. This supports the power of our superpixel approach, taking local patterns as well as spectral responses into consideration.
	
	\subsection{Landscape Species Mapping With SLIC-UAV}
	\label{ssec:dis_land}
	
	The end product of our pipeline is dominance maps for early-successional species (Figure \ref{fig:land_map_BEL_MAC}). We envision that these maps could be used to assess forest condition, according to successional status, and enable recovery to be tracked through time. We see this as a way to view recovery along the dimension of species composition as a proxy for degradation in addition to approaches currently focusing primarily on carbon content, carbon capture and changes in biomass. In addition, this tool could be used to help focus active restoration towards areas towards more degraded areas, where indicative early-successional species are most prevalent, so that assisted natural regeneration techniques such as release cutting, enrichment planting and selective thinning can be implemented \citep{Ansell2011,Swinfield2016}. These approaches have been shown to accelerate carbon sequestration \citep{Reynolds2011,Gourlet2013,Wheeler2016} and the development of suitable habitats for forest specialist species \citep{Ansell2011}. This application for guiding management illustrates the potential value of SLIC-UAV. Our pipeline for species occurrence mapping includes UAVs at all steps, making reference data collection simpler, though care must be taken to ensure species of interest are well represented. We also expect SLIC-UAV to be applicable to high-resolution satellite data. Reference data collection will still need use of UAVs, but the mapping pipeline can be applied to any orthorectified imagery of sufficiently high resolution, where crowns are made up of a at least 100 pixels or so. As satellite imagery continues to increase in resolution this will apply more and more frequently, showing the power of our approach going forward.
	
	We believe the strength of our method on the indicative early-successional species is a particular strength for it's use in operational setting: models focusing on \textit{Macaranga gigantea} and \textit{Bellucia pentamera} performed particularly well, and in moving to the model used for landscape mapping both species were identified with over 90\% precision (Table \ref{table:res_seg_mat}). In Southeast Asian tropical forests, \textit{Macaranga gigantea} is one of several species whose presence is strongly linked to the severity and recency of disturbance, and \textit{Bellucia pentamera} is an invasive species, which is particularly prevalent at Hutan Harapan \citep{Slik2003a,Slik2008}. They are also most commonly found in heavily degraded forest and so can act as a signature species for degradation, or recovery in their absence. We had expected higher prevalence of these species (especially \textit{Bellucia pentamera}) based on our time at the project, but we believe the low occurrence rate is owing to these often growing in the shade of another tree in the upper canopy. Therefore we believe the reported occurrence are for individuals in canopy gaps and we consider these to be the most important cases for management intervention. These individuals are shading out soil without an emergent crown or closed canopy and so are regions particularly worth considering for intervention to accelerate succession. We also note that we only focused on two species, when there are other species we could look at as proxies for disturbance. Including other species would lead to a more complete picture of degradation, as they may live in different microclimate niches from the species we do map, so should be considered when viewing the final maps. We chose species based on perceived and measured occurrence at Hutan Harapan, but consider our work an illustration of the SLIC-UAV pipeline, and the flexibility of the method will allow other species of interest to be added to models with data collection again possible using a UAV, th. Knowledge of the relative rate of occurrence of our chosen species can highlight regions within the project where the effects of disturbance are strongest. It has been shown that structural recovery is quick after fire disturbance, but that this has a longer lasting effect on species composition \citep{Slik2002}. Managers of projects like Hutan Harapan can use approaches like the one we have developed to help distinguish this signature of prior disturbance, based on indicative species occurrence in addition to simpler structural metrics focusing on carbon \citep{Sullivan2017}. This should improve understanding of forest history and current recovery status based on aerial UAV survey, reducing the need to access difficult terrain and the need to use sampling plots to interpolate over management units.

	\subsection{Advantages of full UAV pipeline integration}
	\label{ssec:dis_Adv}
	
	The use of UAVs through the entire pipeline also allows us to reduce the need for mapping by hand in the field, and is lower-cost than arranging for aerial manned aircraft survey. A key step in our work which adds value is our pipeline for collecting training data. Most approaches we have considered have manually mapped crowns in the field with great effort \citep{Lisein2015,Franklin2018,Tuominen2018}, or else have made use of existing forest inventories \citep{Alonzo2018,Fujimoto2019}. In contrast, \citep{Gini2018} work on the UAV imagery, manually labelling individual points on the resulting imagery, but this approach was only pixel-based, using a small proportion of total data for training and testing in an already small study area. \citep{Michez2016}, manually delineated whole crowns on orthomosaic imagery. However, this was completed after processing and not compared to any reference in the field at the time. Our approach improves this and enables collection of training and validation data using a UAV. We set out a clear pipeline for using UAVs to enable generation of reference crown images, with attached GPS location metadata. These can then be used in later digitisations on processed UAV imagery. This not only allows a faster approach to mapping crown locations, but also enables easier access to harder to reach areas. The trade-off for this is focus on only a few key species of restoration interest, enabling production of occurrence maps for these to guide intervention, such as the rule-based approach in \citet{Reis2019}. This process isn\textsc{\char13}t perfect, and will only capture crowns visible from above the canopy, but enables a quick way to construct reference datasets for forest restoration projects, such as at our study site of Hutan Harapan. When focusing on the use of UAVs to guide forest management these should be sufficient reference data, as this approach to management focuses on viewing the canopy from above. This is traded-off with incomplete sampling. We chose to focus on key species which are indicative of degradation and recovery status. However we were able to rapidly build a dataset for species of interest, which we then used to build a mapping model which we were able to apply to 130 ha of data: something that isn\textsc{\char13}t commonly shown by other methods. This shows the power of our approach to aid restoration management. A simple and quickly deployable approach allows collection of data for tree species of interest, which we show can then be used to develop models that can generate heatmaps at the scale of management units (Figure \ref{fig:land_map_BEL_MAC}) to guide restoration. We validate this approach in the structurally complex and biodiverse tropical forest of Hutan Harapan. This is a key ecosystem for global conservation and we have shown the value of our approach in aiding these goals.
	
	\section{Conclusions}
	\label{sec:conclusion}
	
	Our approach to automated species mapping enables a scaling up of local expertise to aid in guiding restoration project managers to monitor and plan interventions. Our SLIC-UAV pipeline centres on the use of UAVs, which are affordable within the budget of most, if not all, projects. The data these can collect are an improvement in both temporal and spatial resolution relative to other remote sensing approaches. We have also shown, within our own pipeline, that it is possible to map upto 100 ha of land a day with a single UAV and operator team of two or three people. This scaling up and deployability of UAV approaches can drastically improve efficiency of human-power in these projects. Our approach is particularly valuable in its ability to map early-successional species of particular management concern. Within Harapan there are already projects exploring the benefit to forest health of selectively logging such species \citep{Swinfield2016}, and our approach will allow more targeted application of such strategies across the landscape. We have also shown how it is possible to collect reference crown location data with UAVs, making use of their high spatial detail combined with GPS information. With local experts it is possible to rapidly collect information on any species of interest, although we focused on four primary species of restoration interest. It would be possible to extend our approach to any species of interest, especially those of high value for biodiversity or for non-timber forest products. This is something which would need validation before deployment, within the same framework we applied. It is important to note that in our modelling, the early-successional species were more visually distinct, thus performance on the species \textit{Alstonia scholaris} and \textit{Endospermum malaccense} are more indicative of expected application to later-successional species. Overall, our work has shown and assessed a full pipeline from field mapping to management occurrence mapping (Figure \ref{fig:land_map_BEL_MAC}) showing the power of this low-cost, easy to implement approach to aid restoration project management.
	
	\section{Acknowledgements}
	
	This project was primarily supported by a NERC CASE studentship partnered with Royal Society for the Protection of Birds (RSPB) [NE/N008952/1] and a grant from the Cambridge Conservation Initiative Collaborative Fund supporting interdisciplinary research between the University of Cambridge and the Royal Society for the Protection of Birds. The field season in 2018 was further supported by a Mark Pryor Grant at Trinity College, Cambridge. We thank Rhett Harrison for his significant input into grant writing. We are grateful to Dr Tuomo Valkonen, whose early attempt to classify species without delineating trees was unsuccessful but paved the way for the development of more sophisticated approaches. We wish to thank all partners at Hutan Harapan for their help with managing the UAV and tree data collection at Hutan Harapan. We particularly wish to thank Adi, Agustiono and Dika for their support with UAV flying and data collection. We are also very grateful for the support from members of Universitas Jambi who supported the logistics of our collaboration. CBS acknowledges support from the Leverhulme Trust project on `Breaking the non-convexity barrier', the Philip Leverhulme Prize, the EPSRC grant [EP/T003553/1], the RISE projects CHiPS and NoMADS, the Cantab Capital Institute for the Mathematics of Information and the Alan Turing Institute. DC was supported by an International Academic Fellowship from the Leverhulme Trust.

	\section{Authors’ contributions}
	JW developed the analysis pipeline, collected the field data with the help of local collaborators, analysed the data and wrote the code, and drafted the paper. BI, EA and MZ guided the conception of the project, developed the plan for the field data collection and facilitated the international collaboration. H and EG oversaw and facilitated the field data collection and supported all work at Hutan Harapan. DAC, CBS and TWS wrote the grant proposals that supported the study, and formed a PhD supervisory team that contributed advice as well as contributing to writing and editing the manuscript.
	
	\singlespacing
	\bibliography{SLICUAV}
	
	\doublespacing
	\section{Supporting Information}
	\label{sec:supI}
	\beginsupplement
	
	\begin{landscape}
		\begin{table}[h]
			\caption{Vegetation and spectral indices computed on a pixel-wise basis. These were based on RGB imagery (RGB), multispectral imagery (MS) or both. Bands are abbreviated as Red (R), Green (G), Blue (B), Red Edge (RE) and Near Infrared (NIR)}
			\begin{center}
				\begin{tabular}{lllll}
					\hline
					Symbol & Name & Imagery & Formula & From \\
					\hline
					{CIG}& {Chlorophyll index} & {MS} & {NIR/G - 1} &  {\citep{Gitelson2003}}\\
					{CVI} & {Cholorphyll vegetation index} &{MS} &  {(NIR$\times$R\textsuperscript{2})}/G & {\citep{Vincini2008}}\\
					{ExR} & {Excess red index} & {RGB}& {2R-G-B} & {\citep{Woebbecke1995}}\\
					{ExG} & {Excess green index} & {RGB} & {2G-R-B} & {\citep{Baron2018}}\\
					{ExB} & {Excess blue index} & {RGB} & {2B-R-G} & {\citep{Baron2018}}\\
					{ExGveg} & {Excess green vegetation index} & {RGB} & {2G-R-B+50} & {\citep{Woebbecke1995}}\\
					{ExRveg} & {Excess red vegetation index} & {RGB} & {(1.4R-G)/(R+G+B)} & {\citep{Meyer2008}}\\
					{ExBveg} & {Excess blue vegetation index} & {RGB} & {(1.4B-G)/(R+G+B)} & {\citep{Mao2003}}\\
					{GLIr} & {Green leaf index - red} & {RGB} & {(2R-G-B)/(2R+G+B)} & {\citep{Goodbody2018b}}\\
					{GLIg} & {Green leaf index - green} & {RGB} & {(2G-R-B)/(2G+R+B)} & {\citep{Louhaichi2001}}\\
					{GLIb} & {Green leaf index - blue} & {RGB} & {(2B-R-G)/(2B+R+G)} & {\citep{Goodbody2018b}}\\
					{GRVI} & {Green-red vegetation index} & {RGB, MS} & {(G-R)/(G+R)} & {\citep{Tucker1979}}\\
					{mGRVI} & {Modified green-red vegetation index} & {RGB, MS} & {(G\textsuperscript{2}-R\textsuperscript{2})/(G\textsuperscript{2}+R\textsuperscript{2})} & {\citep{Bendig2015}}\\
					{IKAW} & {Kawashima index} & {RGB} & {(R-B)/(R+B)} & {\citep{Kawashima1997}}\\
					{NegExR} & {Negative excess red vegetation index} & {RGB, MS} & {G-1.4R} & {\citep{Liu2018}}\\
					{NDVI}& {Normalised difference vegetation index} & {MS} & {(NIR-R)/(NIR+R)} & {\citep{Rouse1974}}\\
					{NDVIg} & {Normalised difference green vegetation index} & {MS} & {(NIR-G)/(NIR+G)} & {\citep{Gitelson1996}}\\
					{NDVIre} &{ Normalised difference red edge vegetation index} & {MS} & {(NIR-RE)/(NIR+RE)} & {\citep{Cavayas2012}} \\
					{RGBVI} & {Red green blue vegetation index} & {RGB} & {(G\textsuperscript{2}-RB)/(G\textsuperscript{2}+RB)} & {\citep{Bendig2015}}\\
					{TGI} & {Triangular greenness index} & {RGB} & {G-0.39R-0.61B} & {\citep{Hunt2011}}\\
					{VARI} & {Visible atmospherically resistant index} & {RGB} & {(G-R)/(G+R-B)} & {\citep{Gitelson2002}}\\
					\hline
				\end{tabular}
			\end{center}
			\label{table:spec_ind}
		\end{table}
	\end{landscape}
	
	\begin{table}[h]
		\caption{Features computed for each spectral band, spectral index and for the DSM heights in each region. Imagery sources are RGB (rgb) top 50\% brightness RGB (top), multispectral (ms), indices (ind), HSV transformed RGB (hsv) and DSM raster (dsm) \textsuperscript{1} excludes DSM imagery as this is based on absolute height AMSL, \textsuperscript{2} applies only to DSM imagery, computing relative heights based on the local median}
		\begin{center}
			\begin{tabular}{lll}
				\hline
				Name & Imagery Applied to & Explanation\\
				\hline
				max\textsuperscript{1} & rgb, top, ms, ind, hsv & maximum pixel value \\
				min\textsuperscript{1}  & rgb, top, ms, ind, hsv & minimum pixel value \\
				mean\textsuperscript{1} & rgb, top, ms, ind, hsv & mean of pixel values \\
				std & rgb, top, ms, ind, hsv, dsm & standard deviation of pixel values \\
				median\textsuperscript{1} & rgb, top, ms, ind, hsv & median of pixel values \\
				cov\textsuperscript{1} & rgb, top, ms, ind, hsv & coefficient of variation of pixel values (mean/std) \\
				skew & rgb, top, ms, ind, hsv, dsm & skewness of pixel values \\
				kurt & rgb, top, ms, ind, hsv, dsm & kurtosis of pixel values \\
				rng & rgb, top, ms, ind, hsv, dsm & range of pixel values \\
				rngsig & rgb, top, ms, ind, hsv, dsm & range expressed in number of standard deviations (rng/std) \\
				rngmean & rgb, top, ms, ind, hsv, dsm & range expressed as multiple of mean value (rng/mean) \\
				deciles\textsuperscript{1} & rgb, top, ms, ind, hsv & 10\textsuperscript{th} to 90\textsuperscript{th} deciles of pixel values \\
				quartiles\textsuperscript{1} & rgb, top, ms, ind, hsv & 1\textsuperscript{st} and 3\textsuperscript{rd} quartiles of pixel values \\
				iqr & rgb, top, ms, ind, hsv, dsm & interquartile range of pixel values \\
				iqrsig & rgb, top, ms, ind, hsv, dsm & IQR expressed in number of standard deviations (iqr/std) \\
				iqrmean\textsuperscript{1} & rgb, top, ms, ind, hsv & IQR expressed as multiple of mean value (iqr/mean) \\
				ratio & rgb, top, ms, ind  & mean of band expressed as fraction of sum of means of all bands \\
				mad\textsuperscript{2} & dsm & median absolute deviation \\
				maxmed\textsuperscript{2} & dsm & maximum value minus median value \\
				minmed\textsuperscript{2} & dsm & minimum value minus median value \\
				decilesmed\textsuperscript{2} & dsm & 10\textsuperscript{th} to 90\textsuperscript{th} deciles of pixel values minus median value\\
				quartilesmed\textsuperscript{2} & dsm & 1\textsuperscript{st} and 3\textsuperscript{rd} quartiles of pixel values minus median value\\
				\hline
			\end{tabular}
		\end{center}
		\label{table:spec_feat}
	\end{table}
	
	\begin{table}[h]
		\caption{Texture features computed for RGB greyscale imagery, multispectral bands and DSM imagery. Class of feature refers to which textural approach this feature is taken from, greylevel co-occurrence matrix (GLCM), local binary pattern (LBP), laws features (LAWS) and autocorrelation (AUTO). \textsuperscript{1} indicates quantised DSM imagery was used, \textsuperscript{2} indicates raw height raster values were used.}
		\begin{center}
			\begin{tabular}{llll}
				\hline
				Name & Class & Offsets & Explanation\\
				\hline
				glcm\_asm\textsuperscript{1} & GLCM & 1,2,3 & mean and range of angular second moment \\
				glcm\_con\textsuperscript{1} & GLCM & 1,2,3 & mean and range of contrast \\
				glcm\_cor\textsuperscript{1} & GLCM & 1,2,3 & mean and range of correlation \\
				glcm\_var\textsuperscript{1} & GLCM & 1,2,3 & mean and range of variance \\
				glcm\_idm\textsuperscript{1} & GLCM & 1,2,3 & mean and range of inverse difference moment \\
				glcm\_sumav\textsuperscript{1} & GLCM & 1,2,3 & mean and range of sum average\\
				glcm\_sumvar\textsuperscript{1} & GLCM & 1,2,3 & mean and range of sum variance\\
				glcm\_sument\textsuperscript{1} & GLCM & 1,2,3 & mean and range of sum entropy \\
				glcm\_ent\textsuperscript{1} & GLCM & 1,2,3 & mean and range of entropy \\
				glcm\_difvar\textsuperscript{1} & GLCM & 1,2,3 & mean and range of difference variance \\
				glcm\_difent\textsuperscript{1} & GLCM & 1,2,3 & mean and range of difference entropy \\
				glcm\_infcor1\textsuperscript{1} & GLCM & 1,2,3 & mean and range of information measure of correlation 1 \\
				glcm\_infcor2\textsuperscript{1} & GLCM & 1,2,3 & mean and range of information measure of correlation 2 \\
				lbp\_1\textsuperscript{1} & LBP & 1 & relative frequencies of the 10 motifs for LBP in a radius of 1 \\
				lbp\_2\textsuperscript{1} & LBP & 2 & relative frequencies of the 18 motifs for LBP in a radius of 1 \\
				lbp\_3\textsuperscript{1} & LBP & 3 & relative frequencies of the 26 motifs for LBP in a radius of 1 \\
				L5E5\textsuperscript{2} & LAWS & N/A & laws feature for detecting intensity of edges \\
				L5S5\textsuperscript{2} & LAWS & N/A & laws feature for detecting intensity of spots \\
				L5R5\textsuperscript{2} & LAWS & N/A & laws feature for detecting intensity of ridges \\
				L5W5\textsuperscript{2} & LAWS & N/A & laws feature for detecting intensity of waves \\
				E5E5\textsuperscript{2} & LAWS & N/A & laws feature for detecting edges in two directions \\
				E5S5\textsuperscript{2} & LAWS & N/A & laws feature for detecting edges occurring perpendicular to spots \\
				E5R5\textsuperscript{2} & LAWS & N/A & laws feature for detecting edges occurring perpendicular to ridges \\
				E5W5\textsuperscript{2} & LAWS & N/A & laws feature for detecting edges occurring perpendicular to waves \\
				S5S5\textsuperscript{2} & LAWS & N/A & laws feature for detecting spots occurring in two directions \\
				S5R5\textsuperscript{2} & LAWS & N/A & laws feature for detecting spots occurring perpendicular to ridges \\
				S5W5\textsuperscript{2} & LAWS & N/A & laws feature for detecting spots occurring perpendicular to waves \\
				R5R5\textsuperscript{2} & LAWS & N/A & laws feature for detecting ridges occurring in two directions \\
				R5W5\textsuperscript{2} & LAWS & N/A & laws feature for detecting ridges occurring perpendicular to waves \\
				W5W5\textsuperscript{2} & LAWS & N/A & laws feature for detecting waves occurring in two directions \\
				acor\_mean\textsuperscript{2} & AUTO & 1,2,3 & mean autocorrelation of image in all directions \\
				acor\_rng\textsuperscript{2} & AUTO & 1,2,3 &  range of autocorrelation across all directions \\
				\hline
			\end{tabular}
		\end{center}
		\label{table:text_feat}
	\end{table}
	
\end{document}